\pdfoutput=1

\documentclass[10pt]{article}

\usepackage[letterpaper,margin=1in]{geometry}
\usepackage{times}
\usepackage[T1]{fontenc}
\usepackage[utf8]{inputenc}
\usepackage{microtype}

\usepackage[round]{natbib}

\usepackage{amsmath}
\usepackage{amssymb}
\usepackage{amsfonts}
\usepackage{bm}
\usepackage{nicefrac}

\usepackage{graphicx}
\usepackage{multirow}
\usepackage{booktabs}
\usepackage{xspace}
\usepackage{caption}
\usepackage{wrapfig}

\newcommand\longmethodname{Decoupled Drift-Diffusion World Model\xspace}
\newcommand\methodname{D3WM\xspace}
\usepackage{amsmath,amsfonts,bm}

\def\eqref#1{equation~\ref{#1}}

\def\1{\bm{1}}

\DeclareMathAlphabet{\mathsfit}{\encodingdefault}{\sfdefault}{m}{sl}
\SetMathAlphabet{\mathsfit}{bold}{\encodingdefault}{\sfdefault}{bx}{n}

\usepackage{url}
\usepackage[colorlinks=true,linkcolor=black,citecolor=black,urlcolor=blue,
            breaklinks=true]{hyperref}
\hypersetup{pdftitle={Decoupled Drift-Diffusion World Models for Consistent Future Frame Prediction},
            pdfauthor={Anton Arapin, Aditya Singh, Yixin Wang, Jun Gao, Bernadette Bucher}}

\title{\bf Decoupled Drift-Diffusion World Models for\\ Consistent Future Frame Prediction}

\usepackage[affil-it]{authblk}

\author[1]{Anton Arapin}
\author[1]{Aditya Singh}
\author[1]{Yixin Wang}
\author[1,2]{Jun Gao}
\author[1]{Bernadette Bucher}
\affil[1]{University of Michigan}
\affil[2]{NVIDIA}

\date{}

\begin{document}

\maketitle
{\let\thefootnote\relax\footnotetext{\hspace*{-1.8em}Correspondence to: Anton Arapin \textless\href{mailto:aarapin@umich.edu}{aarapin@umich.edu}\textgreater\\
Project page: {\urlstyle{same}\url{https://anton-arapin.github.io/d3wm-site/}}}}

\begin{abstract}
World models of visual information predict future image frames conditioned on given action sequences. This predicted scene evolution is driven by deterministic changes such as controlled motion on a robot platform, as well as stochastic changes such as rendering of manipulated objects or previously occluded regions. Our key insight is that the spatial location of these stochastic changes is itself learnable. We leverage this insight by jointly regressing both the deterministic next-state prediction and the spatial uncertainty map, then using the uncertainty map to guide a conditional diffusion process. We propose \longmethodname (\methodname), which uses a transformer trained with a Gaussian negative-log-likelihood objective to produce both the deterministic next-state prediction and a spatial map of heteroscedastic aleatoric uncertainty. The learned uncertainty map then provides a structured spatial gate for a conditional diffusion model: deterministic regions are anchored to the predicted state, while uncertain regions, such as disocclusions or dynamic textures, are sampled by the diffusion process. We frame this decomposition under a variance-gated stochastic differential equation that separates deterministic drift from stochastic diffusion. \methodname shows improved results over competitive baselines in two distinct types of action spaces: robotic tabletop manipulation and camera motion.
\end{abstract}

\begin{figure}[h]
    \centering
     \includegraphics[width=\textwidth]{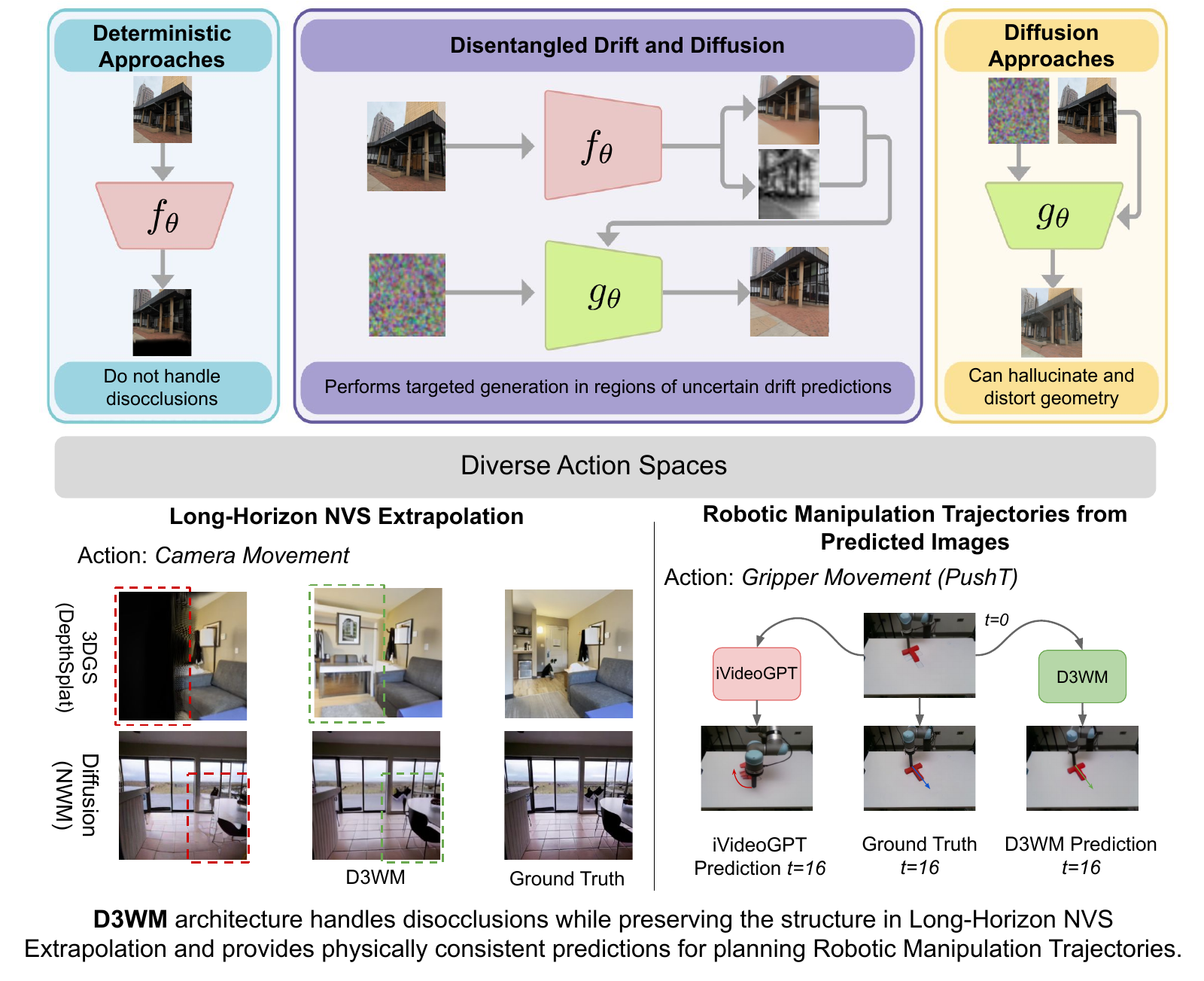}
    \caption{D3WM uses a learnable variance map to disentangle deterministic aspects (drift) from stochastic and texture generation (diffusion) achieving structurally-consistent and photorealistic action-conditioned observation prediction. In contrast, deterministic methods struggle with disocclusions in previously unseen regions, and approaches that rely solely on diffusion are prone to hallucinations and distortions of deterministic parts of the scene.}
    \label{fig:title_figure}
    \vspace{-10pt}
\end{figure}

\section{Introduction}

Perception in humans is an active process in which the outcomes of actions are anticipated before they happen to inform decision making. From throwing a ball to jumping over a puddle, visually grounded outcomes are anticipated based on planned actions~\citep{von1950reafferenzprinzip}. To advance artificial intelligence, computational world models of visual information mimic this biological action-conditioned visual prediction~\citep{zhu2025uwm, zhou2025dinowm, ali2025cosmos}. A natural tension arises in predicting future image frames: how can a model consistently predict deterministic changes from given actions simultaneously with stochastic changes such as rendering of previously occluded regions?

Modern 3D modeling and generative methods address this task by optimizing either the deterministic or stochastic side of the problem at the expense of the other.
Gaussian splats render 3D scenes from image frames, enabling automatic transformations from camera displacements~\citep{charatan23pixelsplat,xu2024depthsplat,chen2024mvsplat}. However, projective transformations alone leave information-less gaps in previously occluded regions.
Alternatively, conditional generative models leverage learned priors to synthesize plausible content for stochastic and deterministic changes~\citep{ren2025gen3c,bar2024navigationworldmodels,wu2024ivideogpt,zhu2025irasim,quevedo2026worldgym}. However, they do not preserve geometric structure in deterministic transitions since these transitions are not explicitly identified.

Recent work bridges these gaps by refining Gaussian splatting novel view synthesis (NVS) results with diffusion to fill information gaps in disoccluded regions~\citep{dong2026one,zhang2025high}. This conditional diffusion approach aligns with our own contributions, but these methods are directly designed to improve performance of NVS where the deterministic transformations are always perspective warps from camera motion.

In our work, we consider predicting future image frames conditioned on arbitrary action sequences — a setting that arises whenever an agent's actions interact with a dynamic environment, as in robotic manipulation. This more general formulation must synthesize the ``world model'' capabilities of generative frameworks with the rigid geometric constraints of 3D modeling. Predicted states must be not only visually plausible but also physically and geometrically consistent with the underlying scene structure and the agent's intended trajectory.

Predicting future frames conditioned on arbitrary actions therefore requires learning the deterministic component of the action's effect, while identifying regions where this deterministic prediction is uncertain. Our key insight is that the spatial location of these uncertain regions is itself learnable: high aleatoric uncertainty in the deterministic prediction identifies areas requiring stochastic generation. We use this spatial uncertainty map to gate generation, retaining geometric structure where prediction is reliable while synthesizing new content where it is not.

\textbf{Contributions.} We introduce \textbf{\longmethodname (\methodname)} to perform action-conditioned prediction of future image frames. \methodname uses a transformer to jointly regress the deterministic next-state prediction and a spatial map of heteroscedastic aleatoric uncertainty. This learned variance map then conditions a diffusion model, guiding the generation of new content toward regions of high aleatoric uncertainty while anchoring the prediction to the deterministic state elsewhere.

We evaluate \methodname on action-conditioned image prediction for robotic manipulation
using the Push-T and BridgeData~V2 datasets. On Push-T, trajectories planned from \methodname's predicted frames are substantially closer to those planned from real frames than with iVideoGPT. On BridgeData~V2, \methodname is the most accurate world model on PSNR at every horizon and the best goal-image candidate verifier, while IRASim remains stronger on policy consistency.
We also validate \methodname on NVS extrapolation from 2-view inputs, extrapolating up to 12 frames into the future on the DL3DV benchmark with the performance gap over the 3DGS baseline widening at longer horizons and demonstrating strong out-of-distribution generalization to RealEstate10k without retraining.

 \section{Related Work}

\textbf{Conditional Generative Models.} Conditional generative architectures enable controllable generation of high-dimensional visual data through explicit structural or physics control~\citep{zhang2023adding,guen2020disentangling}.
Many 3D-aware diffusion pipelines incorporate geometric inductive biases for view synthesis~\citep{chan2023genvs}.
Explicit representations like 3D Gaussian Splatting (3DGS) impose geometric constraints in feed-forward NVS~\citep{charatan23pixelsplat,xu2024depthsplat,chen2024mvsplat,chen2024mvsplat360}.
Recent work reconciles explicit geometry with generative hallucination by jointly optimizing splatting and diffusion for camera-displacement actions~\citep{dong2026one,zhang2025high}.
Similarly, we combine an estimate of the deterministic transformation with targeted diffusion-based refinement. Unlike these methods, we model arbitrary action sequences and predict the deterministic mean alongside a spatial estimate of aleatoric uncertainty, which conditions the diffusion process.

\textbf{SDEs in Neural Networks.}
Neural Stochastic Differential Equations (SDEs) capture continuous-time dynamics by jointly learning deterministic drift and stochastic diffusion components~\citep{pedersen1995consistency, tzen2019neural,liu2020learning,pmlr-v108-li20i,kidger2021efficient}.
However, when applied to high-dimensional data challenges like latent image generation, these approaches experimentally underperform due to the computational intractability of joint pathwise variational inference \citep{course2023amortized} and the failure of isotropic diffusion constraints in highly correlated spatial domains \citep{vandersanden2024edgepreserving}.
Score-based generative models~\citep{song2021scorebased},
flow matching~\citep{lipman2023flow},
and stochastic interpolants~\citep{pmlr-v235-chen24n} bypass joint learning entirely, relying on fixed forward noise processes reversed via score matching.
In contrast, we learn a spatial variance map that controls the location and quantity of novel image features generated, yielding competitive results against large-scale diffusion baselines on standard action-conditioned prediction tasks.

\textbf{World Models for Future Frame Prediction.}
Recent advances in generative modeling have sparked interest in action-conditioned visual world models that simulate complex physical environments.
In the domain of robotics and embodied AI, models such as Navigation World Models (NWM) \citep{bar2024navigationworldmodels}, iVideoGPT \citep{wu2024ivideogpt}, Ctrl-World \citep{guo2025ctrlworld}, and ProphetDWM \citep{prophetdwm2025} have demonstrated that large-scale diffusion and autoregressive architectures can synthesize highly realistic future video frames conditioned on actions.
These models act as neural simulators allowing agents to foresee visual consequences of actions in manipulation or navigation~\citep{hu2025vpp, guo2025ctrlworld,finn2017deep, gao2024vistageneralizabledrivingworld}.

Physical planning in robotics requires geometric consistency to achieve desired outcomes.
Efforts to enforce multi-view and spatial constraints \citep{world-consistent, kim2025dwm, liang2025unifuture, du2026videogpa} have steadily improved the physical plausibility of these generated environments.
However, standard diffusion models map noise to pixel space without disentangling physical state, making them prone to hallucinating structural shifts or overriding action-conditioning in favor of visual priors.

We bridge this gap by explicitly separating deterministic transitions from stochastic generation. Unlike prior works that rely solely on diffusion~\citep{bar2024navigationworldmodels,wu2024ivideogpt,kim2025dwm,world-consistent,zhu2025uwm}, our framework anchors generation to the deterministic outcome of the given actions, using aleatoric uncertainty of the deterministic prediction to gate targeted generation of new scene information. This produces the structural consistency required for reliable trajectory evaluation in downstream policy learning.

\section{Decoupled Drift-Diffusion World Models}

We mathematically formulate a visual prediction architecture built on several intuitive concepts. We first observe that actions yield learnable deterministic changes in a scene. For instance, shifting a camera's viewpoint applies a strict geometric warping to already-visible background structures, just as a robotic actuator command predictably translates a rigid object. Critically, regions requiring hallucination outside of these expected changes can be identified from these deterministic transformations. In visual scenes, previously occluded areas for which there are no prior observations most commonly require this hallucination. Then, the generated information should be predicted using these targeted regions instead of broadly in the scene to preserve deterministic transformations.

We translate these observations into a mathematical formulation of latent future frame prediction. First, we use a deterministic transformer to approximate the predicted mean $\mu_t$ of the deterministic change and aleatoric variance $\sigma_t^2$ in latent space, of the next latent state $z_t$ given a previous latent state $z_0$ and an action sequence $a_{[0,t)}$ for some time horizon $t$ (Section~\ref{subsec:drift}). The predicted variance approximates the spatial heteroscedastic aleatoric uncertainty of the scene~\citep{kendall2017uncertainties}, capturing regions where the deterministic future is inherently ambiguous, such as disocclusions or complex dynamic textures. This variance provides a spatial map for targeted hallucination, used both as conditioning for the diffusion model (Section~\ref{subsec:cond_diff}) and as a spatial gate at inference time (Section~\ref{subsec:inference}).

\subsection{Learning the Predicted Mean and Aleatoric Uncertainty}\label{subsec:drift}

We model the temporal evolution of the latent state under an action sequence as a stochastic process whose deterministic component (the SDE drift coefficient in Eq.~\ref{eq:temporal_sde}) we denote $f_\theta$, and whose stochastic component is characterized by a spatial variance $\sigma_t^2$.
Rather than train $f_\theta$ as an instantaneous rate and integrate it numerically, we directly regress its integrated effect over $[0, t]$, the conditional mean $\mu_t = \mathbb{E}[z_t \mid z_0, a_{[0,t)}]$, together with the aleatoric variance $\sigma_t^2$. The spatial uncertainty $\sigma_t^2$ conditions the diffusion-time refinement of Section~\ref{subsec:inference}.
Throughout the paper we use ``drift'' to refer to the SDE drift coefficient $f_\theta$ in Eq.~\ref{eq:temporal_sde}, while the network's outputs are the predicted mean $\mu_t$ and the variance map $\sigma_t^2$.

We approximate this predicted mean and aleatoric uncertainty using a deterministic transformer $f_\theta^\mu$. Given the previous latent state $z_0$ and action sequence $a_{[0,t)}$, the network outputs the mean and variance parameters of a spatial distribution
\begin{equation}
\mu_t, \sigma_t^2 = f_\theta^\mu(t, z_0, a_{[0,t)}).
\end{equation}
Formally, we model the spatial heteroscedastic aleatoric uncertainty by assuming the ground-truth future state $z_t$ is drawn from a Gaussian distribution parameterized by these outputs, such that we model the probability over spatial locations as a factorized Gaussian: $p(z_t \mid z_0, a_{[0,t)}) = %
\mathcal{N}(z_t; \mu_t, \sigma_t^2 I)$. The network outputs a log-variance squashed by a scaled $\tanh$ into $[-10, 0]$, so $\sigma_t^2 \in [e^{-10}, 1]$ element-wise and the trust map of Eq.~\ref{eq:trust_map} lies in $[0, \lambda_\tau]$.

To ensure $\sigma_t^2$ captures this true aleatoric uncertainty, we train $f_\theta^\mu$ via a negative log-likelihood (NLL) objective~\citep{kendall2017uncertainties}. By treating the target latent state $z_t^{(i,j)}$ at a spatial location $(i,j)$ as a sample from our predicted distribution, the loss becomes:
\begin{equation}
  \mathcal{L}_{\mathrm{drift}} = \frac{1}{2} \sum_{i,j} \left( \log \sigma_t^{2\,(i,j)} + \frac{(z_t^{(i,j)} - \mu_t^{(i,j)})^2}{\sigma_t^{2\,(i,j)}} \right).
\end{equation}

This loss forces the network to predict high variance $\sigma_t^2$ explicitly in spatial regions where the deterministic prediction error is high. We use this variance prediction to create a learned spatial gate for the diffusion process in Section~\ref{subsec:cond_diff}.

\subsection{Conditional Diffusion Formulation}\label{subsec:cond_diff}

We construct a stochastic diffusion model $g_\theta$ using a ControlNet architecture~\citep{zhang2023adding} to sample from a $\mu_t$- and $\sigma_t^2$-conditioned distribution over the next state $\hat{z}_t$. The deterministic prediction network $f_\theta^\mu$ produces the mean $\mu_t$ and variance $\sigma_t^2$ that condition $g_\theta$. To preserve fine textures from the input observation, $g_\theta$ is also conditioned on the reference image frame $z_0$ via IP-Adapter-Plus~\citep{ye2023ip-adapter}. Texture refinement of the predicted mean is performed at inference time via the spatially-gated procedure of Section~\ref{subsec:inference}.

The diffusion model is trained to predict the noise residual $\epsilon$ via the standard conditional $\epsilon$-prediction objective. To avoid notation collision with the temporal frame index $t$, let $\tau \in [0,T]$ denote the diffusion denoising timestep, and let $z_t^{(\tau)}$ denote the noisy target latent at diffusion step $\tau$. We train $g_\theta$ as:
\begin{equation}
   \mathcal{L}_{\mathrm{score}} = \mathbb{E}_{\tau,z_t,\epsilon \sim \mathcal{N}(0, I)} \left[ \| \epsilon - g_{\theta}(z_t^{(\tau)},\tau, \mu_t, \sigma^2_t, z_0) \|^2_2 \right].
\end{equation}
Minimizing this objective is equivalent to minimizing the Fisher Divergence between the data distribution and the model distribution~\citep{song2021scorebased}, learning the conditional gradient field of the next-state distribution given the predicted-mean and variance conditioning.

The deterministic transformer $f_\theta^\mu$ and the conditional diffusion model $g_\theta$ are trained sequentially. At inference, we sample from the learned conditional distribution using the variance-gated procedure detailed in Section~\ref{subsec:inference}, which uses $\sigma_t^2$ as a spatial gate to anchor the diffusion process to the deterministic prediction $\mu_t$ in regions of low aleatoric uncertainty.

\begin{figure}[h!]
    \centering
     \includegraphics[width=\textwidth]{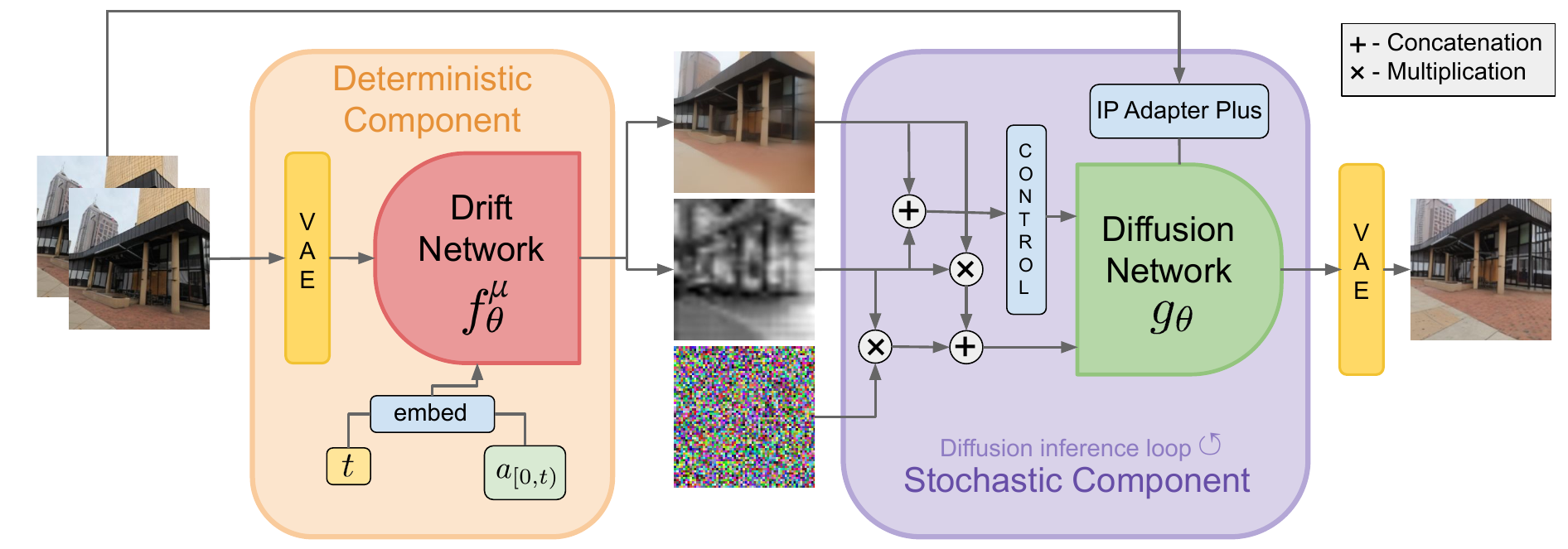}
    \caption{\textbf{Architecture of D3WM.} The input action sequence is denoted $a_{[0,t)}$, and $t$ corresponds to the time shift into the future that the model aims to predict. For a single inference run, our deterministic prediction network $f_\theta^\mu$ runs only once, while the diffusion network $g_\theta$ runs a standard diffusion inference loop using the variance-gated refinement procedure described in Eq.~\ref{eq:inference_refinement}. %
    The deterministic prediction network $f_\theta^\mu$ is labeled `Drift Network' in the figure for compactness, while in the text, ``drift'' refers only to the SDE coefficient $f_\theta$ (Section~\ref{subsec:drift}).}
    \label{fig:drift_net_architecture}
\end{figure}

\subsection{Variance-Gated Inference}\label{subsec:inference}

We formulate the transition to the next future state as a continuous-time stochastic differential equation (SDE) over the observation temporal axis $t$. The dynamics of the scene can be modeled as:
\begin{equation}\label{eq:temporal_sde}
\mathrm{d}z_t = f_{\theta}(t, z_0, a_{[0,t)}) \mathrm{d}t + \sigma_t \odot \mathrm{d} w_t,
\end{equation}
where $f_\theta$ is the deterministic drift coefficient, $\mathrm{d}w_t$ is a standard Wiener process representing physical stochasticity, and $\sigma_t$ is the spatial standard deviation map that scales the uncertainty element-wise. Eq.~\ref{eq:temporal_sde} characterizes the data-generating process, and the network $f_\theta^\mu$ from Section~\ref{subsec:drift} regresses its integrated mean $\mu_t$ and aleatoric variance $\sigma_t^2$ directly rather than numerically integrating $f_\theta$.

A direct numerical integration of this SDE via Euler-Maruyama would sample $w_t$ from an isotropic Gaussian. However, isotropic Gaussian noise produces physically implausible artifacts (e.g., pixelated static in disoccluded regions) rather than realistic textures.
We therefore replace the stochastic term with the trained conditional diffusion model $g_\theta$ from Section~\ref{subsec:cond_diff}, which acts as a learned, image-manifold-aware sampler for the next-state distribution conditional on $\mu_t$ and $\sigma_t^2$.

Concretely, we sample the conditional next-state distribution $p(z_t \mid z_0, a_{[0,t)})$ via the standard Variance-Preserving DDIM reverse process, with a single modification: at each reverse diffusion step we replace the diffusion model's noisy latent with a spatially-weighted combination of the model's prediction and a forward-noised version of the deterministic prediction $\mu_t$. Define the per-pixel latent trust map
\begin{equation}\label{eq:trust_map}
\gamma_t = (\mathbf{1} - \sigma_t^2) \odot \lambda_\tau,
\end{equation}
where $\lambda_\tau \in [0,1]$ is a diffusion-time-dependent anchoring schedule. At reverse step $\tau$, the conditional latent is
\begin{equation}\label{eq:inference_refinement}
\tilde{z}_t^{(\tau)} = \gamma_t \odot \tilde{\mu}_t^{(\tau)} + (\mathbf{1} - \gamma_t) \odot z_t^{(\tau)}, \qquad \tilde{\mu}_t^{(\tau)} = \sqrt{\bar{\alpha}_\tau}\, \mu_t + \sqrt{1 - \bar{\alpha}_\tau}\, \xi_\tau,
\end{equation}
where $\xi_\tau \sim \mathcal{N}(0, I)$ and $\bar{\alpha}_\tau$ is the cumulative noise schedule of the underlying VP diffusion process.
Both terms are valid noisy latents at noise level $\tau$, ensuring $\tilde{z}_t^{(\tau)}$ remains compatible with the next reverse step.
We then proceed with the standard reverse diffusion update using $\tilde{z}_t^{(\tau)}$ in place of $z_t^{(\tau)}$.
The full SDE-level derivation is provided in Appendix~\ref{sec:sde_derivation}.

This formulation enforces our core objective in two limits. In regions of low aleatoric uncertainty ($\sigma_t^2 \to 0$), the trust map $\gamma_t$ approaches $\lambda_\tau$ and the conditional latent is dominated by the noisy deterministic prediction $\tilde{\mu}_t^{(\tau)}$; the final $\hat{z}_t$ collapses toward $\mu_t$, preserving deterministic geometric structure. In regions of high structural ambiguity ($\sigma_t^2 \to 1$), $\gamma_t$ approaches zero and the conditional latent equals the diffusion model's prediction, allowing the diffusion process to synthesize plausible content. The final output of the reverse process at $\tau = 0$ yields our final predicted state $\hat{z}_t$.

\section{Experiments}

\methodname is designed to perform action-conditioned future image frame prediction of stochastic processes. To experimentally validate this capability, we consider two significant types of actions to model: robotic manipulation (Sections~\ref{subsec:robot} and~\ref{sec:bridge}) and camera movement (Section~\ref{subsec:nvs}). In the case of camera movement, we perform 2-view novel view synthesis (NVS) on the DL3DV benchmark~\citep{ling2024dl3dv} and show long horizon extrapolation capabilities of new frames. We also show learning geometric structure can improve generalization of this approach by testing on RealEstate10k~\citep{zhou2018stereo} which we did not train on. For robotic manipulation, we predict image frames after sequences of manipulation actions and use those predicted frames as input to a trajectory generation policy~\citep{chi2024diffusionpolicy,octo2024} trained on the Real PushT dataset~\citep{chi2023diffusionpolicy} and BridgeData~V2 \citep{walke2023bridgedata}.

\subsection{Robotic Manipulation Trajectories from Predicted Images on Real PushT}\label{subsec:robot}

\textbf{Experiment Design.} We design an experiment with robotic manipulation data to evaluate the effectiveness of our method in producing generated images useful for robot control on Real PushT dataset. We use one fixed mounted camera view facing the tabletop and one wrist camera attached to the robot arm as the two input camera views to our network together with trajectories of action sequences to be executed by the robot arm. We predict a frame at the end of this trajectory from each camera view and use it to extend the manipulation planning horizon prior to executing actions for a trained policy by using model's generated frame as input to a trajectory generating policy to generate a new trajectory. We compare the Average Displacement Error (ADE) and Final Displacement Error (FDE) between generated trajectories using a generated image and the ground truth image.

\textbf{Training on Real PushT.} We train \methodname on Real PushT dataset~\citep{chi2023diffusionpolicy,chi2024diffusionpolicy}, with the deterministic network outputting 4 latents per inference: the predicted-mean and variance maps for both the side and wrist camera views. We allow up to 16 actions as conditioning input, where each action vector concatenates gripper world pose, arm rotation, and time shift.

\textbf{Evaluation on Real PushT.} When designing our prediction architecture, we claimed that the physical and geometric consistency our model enforces is critical to successful use of generated images in robotic control. To test this, we select 100 frames at random from each of the 14 episodes in the test split of the Real PushT dataset. For each of these frames, we run D3WM to predict 16 frames into the future conditioned on the corresponding actions. We then use our predicted frames as input to the pre-trained PushT Diffusion Policy~\citep{chi2023diffusionpolicy,chi2024diffusionpolicy} to generate a trajectory. We do this setup on the same randomly selected frames using iVideoGPT~\citep{wu2024ivideogpt}, which was trained on a subset of Open X-Embodiment~\citep{open_x_embodiment_rt_x_2023} dataset including the Real PushT dataset. We compute the Average Displacement Error (ADE) and Final Displacement Error (FDE) between trajectories and the trajectory generated by running PushT Diffusion Policy on the ground truth image.

\begin{figure}[htbp]
    \centering

    \begin{minipage}[b]{0.35\textwidth}
        \centering
        \setlength{\tabcolsep}{3pt}
        \begin{tabular}{@{}lcc@{}}
            \toprule
            Method & ADE $\downarrow$ & FDE $\downarrow$ \\
            \midrule
            Ground Truth & 1.552 & 1.319 \\
            iVideoGPT    & 45.134 & 65.228 \\
            \textbf{D3WM}         & \textbf{12.864} & \textbf{17.550} \\
            \bottomrule
        \end{tabular}

        \makeatletter\def\@captype{table}\makeatother
        \caption{Quantitative result comparing ADE and FDE of the manipulation trajectories generated from image predicted by D3WM and iVideoGPT. "Ground Truth" is the average ADE and FDE error from the stochastic generation of the trajectory when run twice in a row on the same ground truth frame.}
        \label{tab:pusht_quant}
    \end{minipage}
    \hfill
    \begin{minipage}[b]{0.62\textwidth}
        \centering

        \makebox[\linewidth][c]{\includegraphics[width=1.07\linewidth]{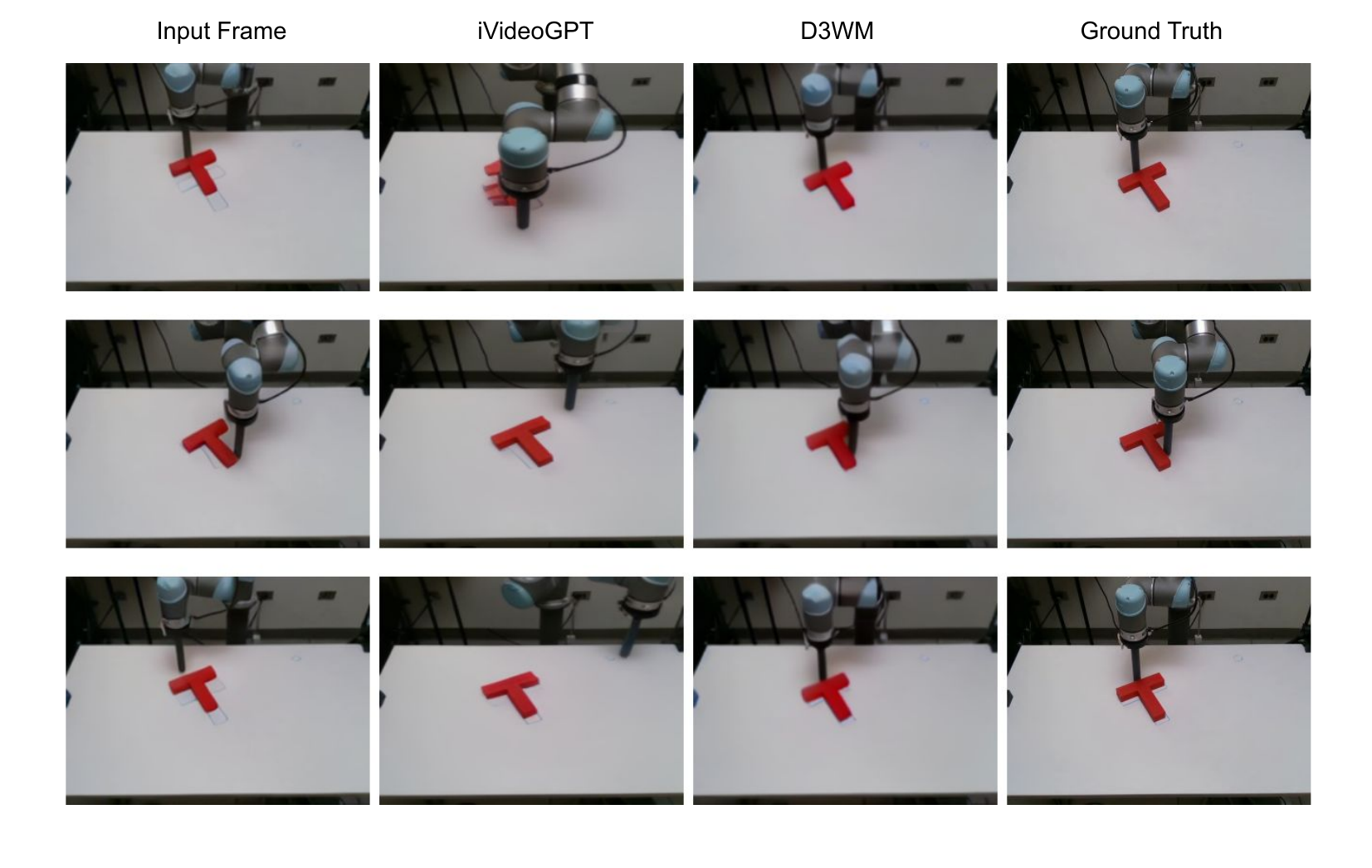}}

        \makeatletter\def\@captype{figure}\makeatother
        \caption{Qualitative comparison of predicting 16 frames into the future on Real PushT dataset. D3WM achieves high accuracy in terms of the gripper and "T"-object placement, while iVideoGPT suffers from hallucinated gripper movement and object deformations.}
        \label{fig:pusht_qual}
    \end{minipage}

\end{figure}

\textbf{Trajectory Generation Results.} As shown in Table \ref{tab:pusht_quant}, we significantly outperform iVideoGPT on trajectory error on both ADE and FDE. This result highlights that our predictions are much more useful for robotic planning and control. The qualitative results in Figure~\ref{fig:pusht_qual} show that iVideoGPT distorts and hallucinates geometry, resulting in physically impossible gripper and object locations as well as distortions of the rigid T object being pushed. In contrast, \methodname accurately models deterministic changes including physically consistent placements of the gripper and object.

\subsection{BridgeData~V2}
\label{sec:bridge}

We evaluate D3WM on BridgeData~V2 \citep{walke2023bridgedata}, which is a dataset of tabletop manipulation by a WidowX 250 robotic arm, with fixed camera observations and 7-dimensional actions (6-DoF end-effector motion plus gripper). We compare against two generative world models for this dataset, IRASim \citep{zhu2025irasim} and WorldGym \citep{quevedo2026worldgym}.

\begin{figure}[t]
\centering
\includegraphics[width=\textwidth]{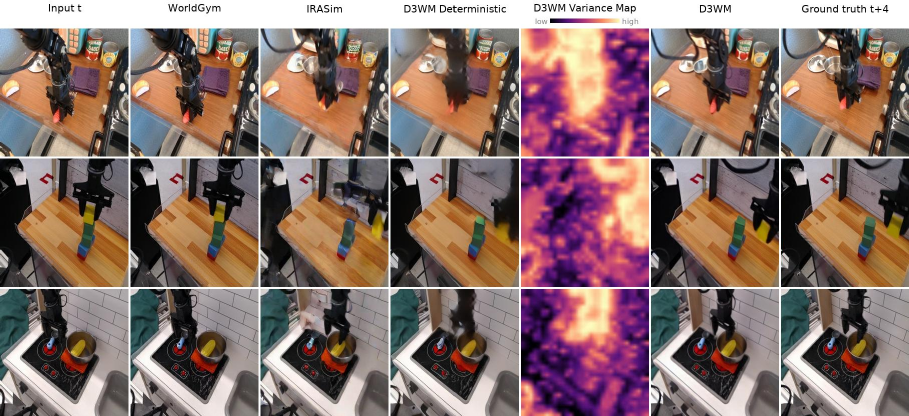}
\caption{Qualitative results on BridgeData~V2. The D3WM deterministic prediction hedges and blurs, which is what earns it the pixel metrics in Table~\ref{tab:bridge_rollout}, while the D3WM full model samples a sharp completion and is scored against the single future that model committed to. WorldGym under-commands and IRASim shows occasional hallucinations or wrong motion.}
\label{fig:bridge_qual}
\end{figure}

\textbf{Open-loop rollouts.} Table~\ref{tab:bridge_rollout} shows both D3WM deterministic-only ($\mu_t$) and full D3WM variants taking the top two positions among world models on PSNR,
with D3WM $\mu_t$'s lead over the strongest other world model growing with the horizon, from $+0.9$\,dB at $g{=}1$ to $+1.6$\,dB at $g{=}4$ ($+1.2$ to $+1.6$\,dB over IRASim),
and up to $+3.6$\,dB over the copy-frame control, with the full model second throughout. On LPIPS, D3WM $\mu_t$ leads every model from $g{=}3$ and is the only one to pass the copy control at $g{=}4$ ($0.112$ against $0.121$). %
D3WM $\mu_t$ produces mean blur which is favored by PSNR, as shown in Fig.~\ref{fig:bridge_qual}, while the full D3WM model, in contrast, samples sharp outcomes. Whether the sharpness is worth what it costs in PSNR is not decidable from Table~\ref{tab:bridge_rollout} alone.

\textbf{Goal-image candidate verification.}
At each state, $K$ candidate action chunks are ranked by
predicted latent similarity to the true $t{+}4$ frame, and the selected chunk is scored against the demonstrated action. Random-of-$K$ and oracle-of-$K$
bracket every row, so a verifier is read by the fraction of that gap it
captures (Table~\ref{tab:budget}). D3WM  is the best verifier in
the table, beating IRASim at every pool size, $38$--$46\%$ of the
random-to-oracle gap closed against $23$--$26\%$, at half the compute
per candidate. WorldGym gets worse as the pool grows ($0.044 \to
0.063$), and it under-commands, so candidates that command little motion are the ones it scores best. The D3WM $\mu_t$ row is weak because an averaged prediction is close to every candidate's outcome and discriminates between none of them, though at $4.9$\,ms per candidate it remains the only variant cheap enough to screen at control rate.

\textbf{Policy consistency.} We measure ADE and FDE as in the
real Push-T experiment (\S\ref{subsec:robot}), using the Octo policy \citep{octo2024}, except that the policy acts on the predicted and the real future under paired sampling keys, so a perfect simulator scores $0$. Table~\ref{tab:bridge_ade}
shows that the stochastic component is worth a large margin over the deterministic alone ($1.063$ against $1.445$ ADE), which is the same sharpness that cost it PSNR now being paid back by a policy that was never trained on blurred input. D3WM (full) is second behind IRASim, and the gap is structural
rather than a deficit of fidelity, since IRASim diffuses the whole trajectory
and can enforce consistency across the horizon, where D3WM, like
WorldGym, predicts a single frame and has no mechanism tying
independently predicted futures together, traded-off by its $2\times$ speedup compared to IRASim.

\begin{table}[t]\centering\small
\begin{tabular}{l cccc cccc}
\toprule
& \multicolumn{4}{c}{PSNR$\uparrow$} & \multicolumn{4}{c}{LPIPS$\downarrow$}\\
\cmidrule(lr){2-5}\cmidrule(lr){6-9}
gap $g$ & 1 & 2 & 3 & 4 & 1 & 2 & 3 & 4\\
\midrule
copy-frame (control) & 24.48 & 21.48 & 20.04 & 19.19 & \textbf{.051} & \textbf{.084} & \textbf{.105} & .121\\
IRASim & 23.34 & 22.37 & 21.64 & 21.12 & .101 & .116 & .127 & .136\\
WorldGym & 23.80 & 21.10 & 19.70 & 18.86 & .059 & .093 & .118 & .137\\
D3WM $\mu_t$ & \textbf{24.73} & \textbf{23.54} & \textbf{23.09} & \textbf{22.76} & .086 & .099 & .107 & \textbf{.112}\\
D3WM & 24.29 & 22.80 & 22.32 & 21.91 & .104 & .118 & .124 & .130\\
\bottomrule
\end{tabular}
\caption{Open-loop rollout quality on BridgeData~V2. The two
D3WM variants hold the top two model positions on
PSNR at every gap, and the D3WM $\mu_t$ lead over the best baseline world model grows with the horizon. Both metrics here are
reference-based and therefore reward a hedged prediction, where the D3WM deterministic prediction leads partly because it blurs, and the full D3WM model, which samples a sharp completion, is penalized for missing the one future that occurred (Fig.~\ref{fig:bridge_qual}). The copy-frame control wins on LPIPS because scenes are dominated by the mostly static background, passed only at $g{=}4$ by D3WM deterministic prediction.}
\label{tab:bridge_rollout}
\end{table}

\begin{table}[t]
\centering
\small
\begin{minipage}[t]{0.56\linewidth}
  \centering
  \setlength{\tabcolsep}{3pt}
  \begin{tabular}{l r r r r r}
    \toprule
    Verifier & ms/cand & $K$@1s & \multicolumn{3}{c}{selected-action MSE at $K$} \\
    \cmidrule(lr){4-6}
    & & & 4 & 16 & 32 \\
    \midrule
    random & --- & --- & .040 & .038 & .038 \\
    oracle & --- & --- & .006 & .003 & .003 \\
    IRASim & 5478 & 0.2 & .031 & .029 & .030 \\
    D3WM $\mu_t$ & \textbf{4.9} & \textbf{204} & .054 & .043 & .038 \\
    D3WM  & 2643 & 0.4 & \textbf{.027} & \textbf{.024} & \textbf{.022} \\
    WorldGym & 1094 & 0.9 & .044 & .056 & .063 \\
    \bottomrule
  \end{tabular}
  \caption{Goal-image candidate verification. Each world
  model ranks $K$ candidate action chunks by predicted similarity to the
  true $t{+}4$ frame, and the selected chunk is scored against the
  demonstrated action. Random-of-$K$ and oracle-of-$K$ bracket every
  row. The D3WM deterministic prediction discriminates weakly due to mean blur, and D3WM full method wins due to sharp and accurate sampling.}
  \label{tab:budget}
\end{minipage}%
\hfill
\begin{minipage}[t]{0.41\linewidth}
  \centering
  \setlength{\tabcolsep}{4pt}
  \begin{tabular}{lcc}
    \toprule
    world model & ADE$\downarrow$ & FDE$\downarrow$\\
    \midrule
    copy-frame (control) & 1.577 & 1.994\\
    WorldGym & 1.448 & 1.783\\
    D3WM $\mu_t$ & 1.445 & 1.649\\
    D3WM & 1.063 & 1.193\\
    IRASim & \textbf{0.846} & \textbf{1.035}\\
    \bottomrule
  \end{tabular}
  \caption{Policy consistency. Frozen Octo policy acts on the predicted and
  on the real future under paired sampling keys, so a perfect simulator
  scores $0$. The D3WM full method is worth a large margin over D3WM deterministic-only prediction, but IRASim leads since it diffuses the whole trajectory and can enforce consistency across the horizon, where D3WM and WorldGym predict single frames.}
  \label{tab:bridge_ade}
\end{minipage}
\end{table}

\subsection{NVS Extrapolation}\label{subsec:nvs}

\textbf{DL3DV Benchmark Experiment Design.} We perform a two-view long-horizon NVS extrapolation task, where we seek to predict an observation conditioned on the action of camera movement. We train our model with 2-view input, conditioned on the action of camera movement on DL3DV dataset~\citep{ling2024dl3dv}. We include camera intrinsics together with the input views, matching the formulation in NVS baseline methods. %
We evaluate on a subset of DL3DV Benchmark, predicting the observation 3, 6, 9 and 12 frames into the future. The DL3DV Benchmark contains 140 episodes, with hundreds of frames each. For our subset we randomly sample 20 frames and associated poses for each episode, for each time shift 3, 6, 9 and 12. We compare the performance of D3WM to the contemporary 3DGS state-of-the-art approach DepthSplat~\citep{xu2024depthsplat}, using the same two reference frames.

\begin{table}[h]
    \centering
    \setlength{\tabcolsep}{1.85pt} %

    \begin{minipage}[t]{0.45\textwidth}
        \centering
        \begin{tabular}{llccc}
            \toprule
            Extrapol. & Method & PSNR $\uparrow$ & SSIM $\uparrow$ & LPIPS $\downarrow$ \\
            \midrule
            \multirow{2}{*}{3 fr.}
            & DepthSplat    & 20.8678 & \textbf{0.7610} & \textbf{0.1997} \\
            & D3WM & \textbf{21.4845} & 0.6690 & 0.2218 \\
            \midrule
            \multirow{2}{*}{6 fr.}
            & DepthSplat    & 17.3864 & \textbf{0.6431} & 0.2975 \\
            & D3WM & \textbf{20.1228} & 0.6092 & \textbf{0.2829} \\
            \midrule
            \multirow{2}{*}{9 fr.}
            & DepthSplat    & 15.3978 & 0.5565 & 0.3706 \\
            & D3WM & \textbf{19.1892} & \textbf{0.5629} & \textbf{0.3378} \\
            \midrule
            \multirow{2}{*}{12 fr.}
            & DepthSplat    & 14.1356 & 0.4955 & 0.4208 \\
            & D3WM & \textbf{18.3521} & \textbf{0.5245} & \textbf{0.3824} \\
            \bottomrule
        \end{tabular}
        \caption{Comparison to DepthSplat on DL3DV in-distribution Benchmark.}
        \label{tab:nvs_depthsplat_quant}
    \end{minipage}
    \hfill %
    \begin{minipage}[t]{0.45\textwidth}
        \centering
        \begin{tabular}{llccc}
            \toprule
            Extrapol. & Method & PSNR $\uparrow$ & SSIM $\uparrow$ & LPIPS $\downarrow$ \\
            \midrule
            \multirow{2}{*}{3 fr.}
            & NWM & 18.4498 & 0.6318 & 0.2334 \\
            & D3WM & \textbf{21.5518} & \textbf{0.7901} & \textbf{0.1415} \\
            \midrule
            \multirow{2}{*}{6 fr.}
            & NWM & 15.8612 & 0.5541 & 0.3145 \\
            & D3WM & \textbf{19.2070} & \textbf{0.6191} & \textbf{0.1891} \\
            \midrule
            \multirow{2}{*}{9 fr.}
            & NWM & 14.4502 & 0.5086 & 0.3750 \\
            & D3WM & \textbf{17.6283} & \textbf{0.5561} & \textbf{0.2369} \\
            \midrule
            \multirow{2}{*}{12 fr.}
            & NWM & 13.5672 & 0.4795 & 0.4219 \\
            & D3WM & \textbf{16.7798} & \textbf{0.5223} & \textbf{0.2658} \\
            \bottomrule
        \end{tabular}
        \caption{Comparison to NWM on out-of-distribution RealEstate10K.}
        \label{tab:nvs_nwm_quant}
    \end{minipage}
    \vspace{-10pt}
\end{table}

\textbf{Ablations.} We ablate the components of D3WM at $t=3,6,9,12$ frame extrapolation on the DL3DV Benchmark dataset (Figure~\ref{fig:ablations_plot}). The \textit{Deterministic Only} baseline ($\mu_t$ alone) achieves the highest PSNR/SSIM by producing locally-blurred predictions in disocclusion regions, but also the second-worst LPIPS (only \textit{Gate Only} is worse). PSNR rewards averaging over plausible outcomes since per-pixel deviation remains bounded, while LPIPS penalizes the lack of high-frequency structural detail this averaging produces. Diffusion-based variants commit to specific high-frequency textures, improving perceptual realism at the cost of pixel-wise deviation when committing incorrectly. We evaluate on both metrics throughout. Variance gating ablations are discussed in App.~\S\ref{sec:full_ablation_table}.

\begin{figure}[h!]
    \centering
     \includegraphics[width=\textwidth]{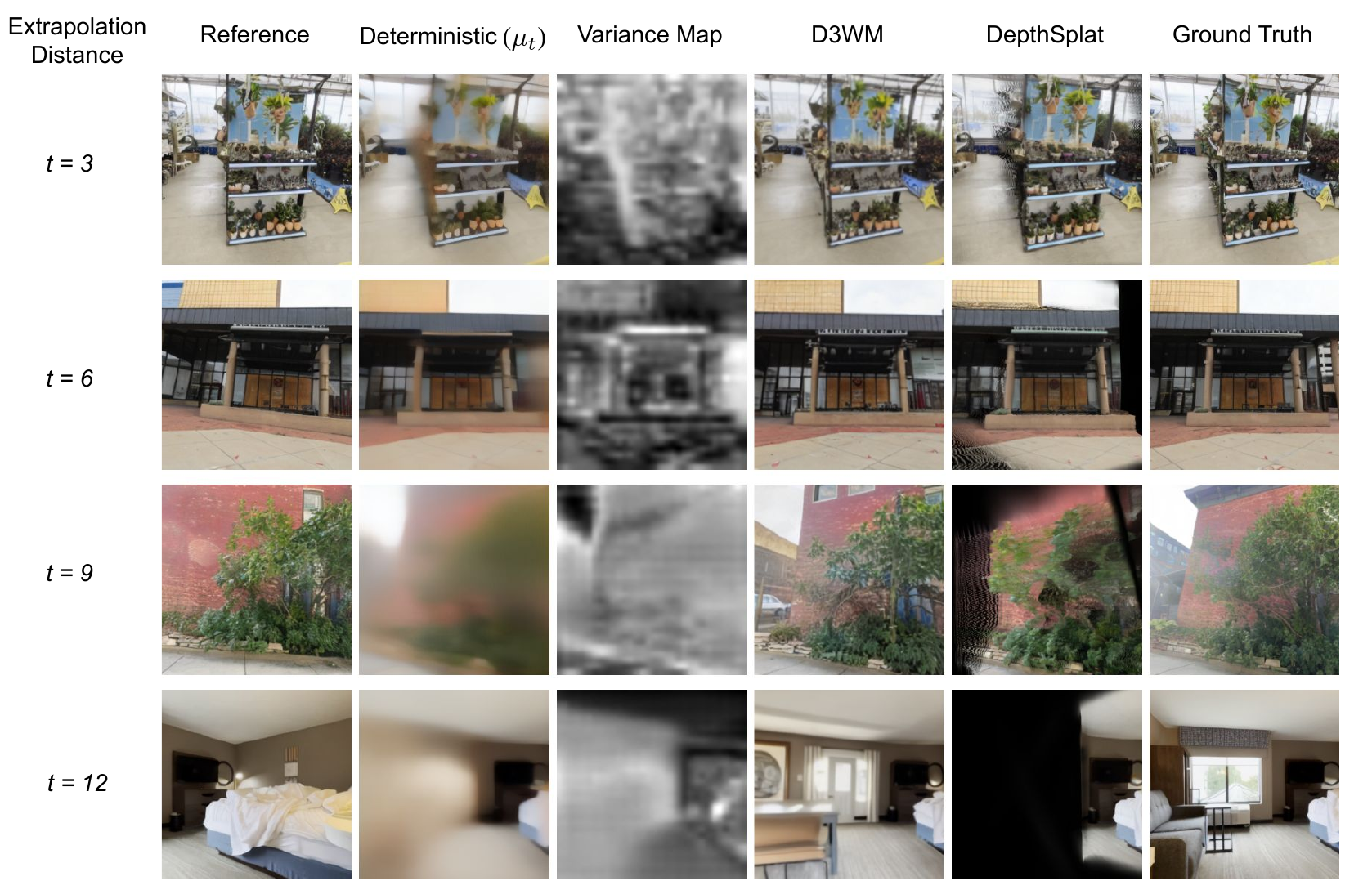}
    \caption{Qualitative comparison of 2-view NVS extrapolation between \methodname and DepthSplat, predicting 3, 6, 9, and 12 frames into the future. \methodname identifies regions of high variance (Variance Map) and uses the diffusion network to reconstruct textures and generate previously occluded regions, while preserving the geometric correctness provided by the deterministic prediction. DepthSplat produces ``black holes'' in regions that were previously occluded.}
    \label{fig:nvs_depthsplat_qual4row}
    \vspace{-5pt}
\end{figure}

\textbf{DL3DV Benchmark Results.} As shown in Table~\ref{tab:nvs_depthsplat_quant}, D3WM outperforms DepthSplat at longer extrapolation horizons. We outperform on PSNR at every horizon, on LPIPS for 6 frames or further, and on all three metrics (PSNR, SSIM, LPIPS) at 9 and 12 frames. The performance gap widens with horizon: DepthSplat's reliance on warping pixels from the input views breaks down as the camera moves further from the input viewpoint, leaving expanding ``black holes'' in disoccluded regions where no input pixels can be projected (Figure~\ref{fig:nvs_depthsplat_qual4row}). D3WM's diffusion network handles these regions cleanly, while the variance-gated bridge anchors the prediction to the deterministic projection elsewhere. The remaining SSIM gap at short horizons (3 and 6 frames), and the LPIPS gap at 3 frames, reflect the same Gaussian-splatting advantage observed in our ablations: in regions where geometric warping is sufficient, splatting derives pixels directly from input frames, producing very high SSIM scores. D3WM's diffusion-synthesized textures look natural but deviate at the pixel level.

\textbf{Out-of-Distribution Generalization to RealEstate10k.}
We also perform a 2-view long-horizon NVS extrapolation experiment on an out-of-distribution dataset. We use a subset of RealEstate10K dataset~\citep{zhou2018stereo} for evaluation, from which we select 80 episodes and randomly sample 20 frames and associated poses for each episode. We compare our performance to that of Navigation World Models (NWM)~\citep{bar2024navigationworldmodels} on the same out-of-distribution evaluation dataset, using the same reference and target frames. Our method was only trained on data from DL3DV.

We outperform NWM on out-of-distribution NVS extrapolation in terms of PSNR, SSIM and LPIPS for all time shifts (3, 6, 9 and 12), as shown in Table \ref{tab:nvs_nwm_quant}. Although diffusion approaches like NWM are very good at high-resolution details, they often distort structure and hallucinate artifacts, as can be seen in our qualitative results in Appendix~\ref{sec:nvs_discussion}. The structural anchoring provided by D3WM's variance-gated diffusion process produces both higher PSNR and lower LPIPS than NWM's free-form diffusion approach, demonstrating that the architectural separation of deterministic dynamics from stochastic generation generalizes beyond the training distribution.

\section{Conclusion}

We introduce \longmethodname (\methodname), a novel method for action-conditioned future frame prediction. \methodname jointly regresses the deterministic next-state prediction and a spatial map of heteroscedastic aleatoric uncertainty, then uses this uncertainty map to gate generation by a conditional diffusion model. We validate \methodname on novel view synthesis with camera-motion actions and
on robotic manipulation prediction. On NVS, \methodname outperforms DepthSplat at longer horizons and NWM on out-of-distribution RealEstate10k, and on manipulation, its predictions improve downstream trajectory planning on Push-T and goal-image candidate verification on BridgeData~V2.

\textbf{Limitations and Future Work.} While \methodname can predict arbitrary horizons into the future, it produces only a single frame at each query. This is efficient when only the trajectory endpoint matters (e.g., the result of a single robotic manipulation), but limiting when assessing the viability of intermediate states. Our formulation also assumes scene changes are driven entirely by the input action sequence, and does not account for predictable agent-independent dynamics. Finally, while we input two reference views, the architecture is not restricted to two, hence we plan to experiment with varying view counts and to extend the targeted-generation framework beyond RGB images.

\subsection*{Ethics statement}
This work trains generative world models on public datasets of robot manipulation and real-world scenes. It involves no human subjects, no personally identifying data and no new data collection. Predicted frames can look plausible while being wrong, so they should not stand in for real observations in safety-critical control without validation.

\subsection*{Reproducibility statement}
The Method section and Appendix provide the full architectural and evaluation details.

\subsection*{AI use statement}
Generative AI tools were used for manuscript drafting and copy-editing, for producing explanatory diagrams from the authors' own figure specifications, and for
searching and verifying related-work citations, as well as polishing code for the release. They were not used for research ideation, experimental design, or analysis. Models, experiments and all reported numbers are the authors' own work. The authors reviewed all AI-assisted text and figures, verified every number against the
experiment logs that produced it, and take responsibility for the final content of this work.

\bibliography{main}
\bibliographystyle{iclr2027_conference}

\appendix
\section{Expanded Ablation Results}
\label{sec:full_ablation_table}

\begin{figure}[h!]
     \centering
      \includegraphics[width=\textwidth]{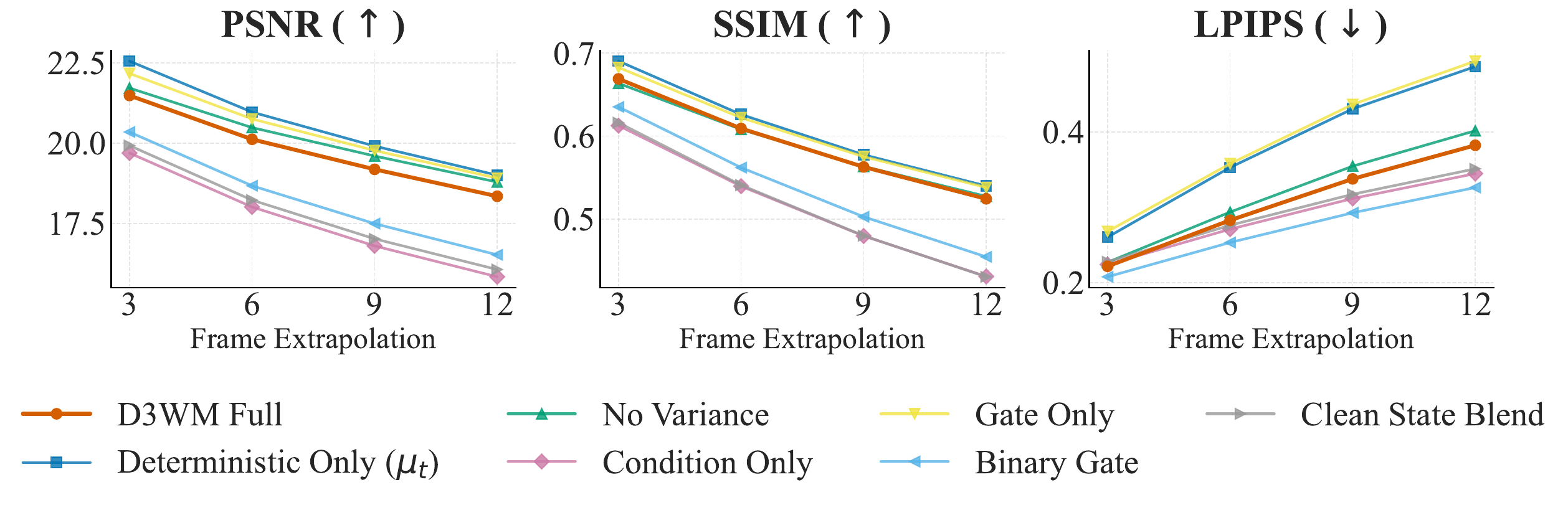}
     \caption{Ablations of D3WM method. The \textit{D3WM Full} provides the best balance of structural similarity, signal fidelity and perceptual realism.}
     \label{fig:ablations_plot}
     \vspace{-5pt}
 \end{figure}

\textbf{Variance Gate Ablations.} Here we describe the full list of ablations shown in Figure~\ref{fig:ablations_plot} and Table \ref{tab:nvs_ablations_quant}. Disabling all variance-aware components (\textit{No Variance}) recovers a standard ControlNet baseline that improves LPIPS over \textit{Deterministic Only} but loses structural fidelity. Variance used only as ControlNet conditioning without inference-time gating (\textit{Condition Only}) underperforms both on PSNR and SSIM, indicating that variance is most useful as an inference-time spatial gate rather than as a training-time conditioning signal alone. The D3WM Full method, combining
variance conditioning, the variance attention bias (Appendix~\ref{sec:arch}), and the spatially-gated bridge
of Eq.~\ref{eq:inference_refinement}, improves LPIPS over \textit{Deterministic Only}, \textit{Gate Only} and \textit{No Variance} at every horizon while staying within 1.1\,dB PSNR of \textit{Deterministic Only}. Variance gating without ControlNet conditioning
(\textit{Gate Only}) stays within 0.4\,dB PSNR of \textit{Deterministic Only} but has the worst LPIPS at every horizon; adding the variance-conditioned ControlNet and attention bias (\textit{D3WM Full}) improves LPIPS by 0.05--0.11 at a cost of 0.6--0.7\,dB PSNR. The inference-time gate therefore carries the structural fidelity, and the conditioning supplies the perceptual detail.
Binarizing the variance map (\textit{Binary Gate}) improves LPIPS at the cost of structural fidelity, suggesting the continuous variance values are calibrated meaningfully rather than just as a hard occlusion mask. Replacing the noisy-bridge gate of Eq.~\ref{eq:inference_refinement} with a clean-state posterior-mean formulation (\textit{Clean State Blend}) drops PSNR by approximately 1.97 dB averaged across extrapolation horizons and reverts to the \textit{Condition Only} regime, confirming that the specific choice of where to apply the variance gate during the reverse diffusion process is empirically consequential. %

\begin{table}[t]
  \centering
  \begin{tabular}{llccc}
    \toprule
    Extrapolation & Ablation & PSNR $\uparrow$ & SSIM $\uparrow$ & LPIPS $\downarrow$ \\
    \midrule
    \multirow{7}{*}{3 frames}
    & \textbf{D3WM Full}    & 21.4845 & 0.6690 & 0.2218 \\
    & Deterministic Only ($\mu_t$)    & 22.5551 & 0.6904 & 0.2605 \\
    & Condition Only & 19.6972 & 0.6127 & 0.2242 \\
    & Gate Only & 22.1707 & 0.6827 & 0.2680 \\
    & Binary Gate & 20.3500 & 0.6352 & 0.2079 \\
    & Clean State Blend & 19.9284 & 0.6162 & 0.2279 \\
    \midrule
    \multirow{7}{*}{6 frames}
    & \textbf{D3WM Full}    & 20.1228 & 0.6092 & 0.2829 \\
    & Deterministic Only ($\mu_t$)    & 20.9693 & 0.6261 & 0.3530 \\
    & Condition Only & 18.0184 & 0.5397 & 0.2711 \\
    & Gate Only & 20.7563 & 0.6222 & 0.3580 \\
    & Binary Gate & 18.6787 & 0.5621 & 0.2534 \\
    & Clean State Blend & 18.2324 & 0.5410 & 0.2763 \\
    \midrule
    \multirow{7}{*}{9 frames}
    & \textbf{D3WM Full}    & 19.1892 & 0.5629 & 0.3378 \\
    & Deterministic Only ($\mu_t$)    & 19.9124 & 0.5776 & 0.4307 \\
    & Condition Only & 16.8081 & 0.4798 & 0.3118 \\
    & Gate Only & 19.7741 & 0.5753 & 0.4366 \\
    & Binary Gate & 17.5000 & 0.5029 & 0.2928 \\
    & Clean State Blend & 17.0277 & 0.4801 & 0.3174 \\
    \midrule
    \multirow{7}{*}{12 frames}
    & \textbf{D3WM Full}    & 18.3521 & 0.5245 & 0.3824 \\
    & Deterministic Only ($\mu_t$)    & 19.0078 & 0.5400 & 0.4868 \\
    & Condition Only & 15.8471 & 0.4310 & 0.3448 \\
    & Gate Only & 18.9141 & 0.5381 & 0.4941 \\
    & Binary Gate & 16.5306 & 0.4546 & 0.3263 \\
    & Clean State Blend & 16.0728 & 0.4306 & 0.3512 \\
    \bottomrule
  \end{tabular}
  \caption{Ablations of D3WM method. The \textit{D3WM Full} provides the best balance of structural similarity, signal fidelity and perceptual realism.}
  \label{tab:nvs_ablations_quant}
  \vspace{-20pt} %
\end{table}

\textbf{Schedule Variants.} In our final configuration we use a
step-function schedule for $\lambda_\tau$ in Eq.~\ref{eq:trust_map} that disables anchoring after the first 70\% of denoising steps. We compared this against a quarter-cosine schedule (smooth decay over all steps), linear ramp, and constant anchoring across all steps. The step function performed comparably to the cosine schedule and outperformed both the linear and constant variants. The constant schedule preserved structural fidelity (high SSIM) but produced blurry textures (worse LPIPS), indicating that the diffusion model needs unanchored final steps to develop coherent texture detail.
The linear ramp was the worst of the four.
We do not include these variants in the main paper as their effect was small relative to the design choices reported in Figure~\ref{fig:ablations_plot}.

\textbf{Variance Map Shape Variants.} We also investigated whether
modifying the dynamic range of the variance map changes downstream quality. Raising the variance map to power $0.5$ before use widens high-uncertainty regions (favoring more diffusion-driven generation), while raising to power $2.0$ shrinks them. Both modifications produced small changes to metrics consistent with the corresponding shift in the trust threshold, suggesting the learned variance values are calibrated meaningfully and robust to monotonic transformations.

\begin{figure}[h!]
    \centering
     \includegraphics[width=\textwidth]{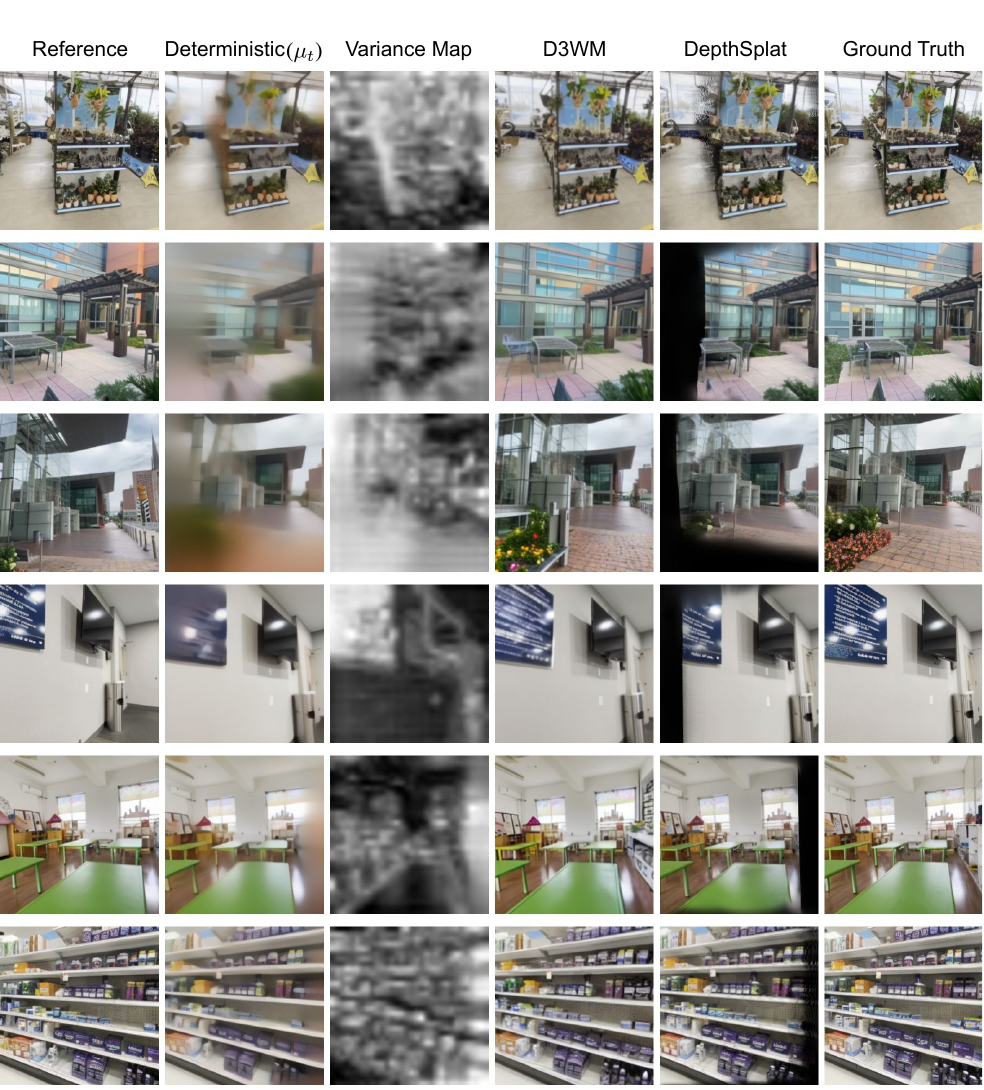}
    \caption{Additional results for qualitative comparison between D3WM and DepthSplat on DL3DV Benchmark dataset with extrapolation time shift $t=3$ frames into the future. Highlighting how D3WM handles the "black hole" problem as compared to 3DGS approaches while retaining structural consistency of the rest of scene.}
    \label{fig:dsdisp3_qual_supplement1}
\end{figure}

\begin{figure}[h!]
    \centering
     \includegraphics[width=\textwidth]{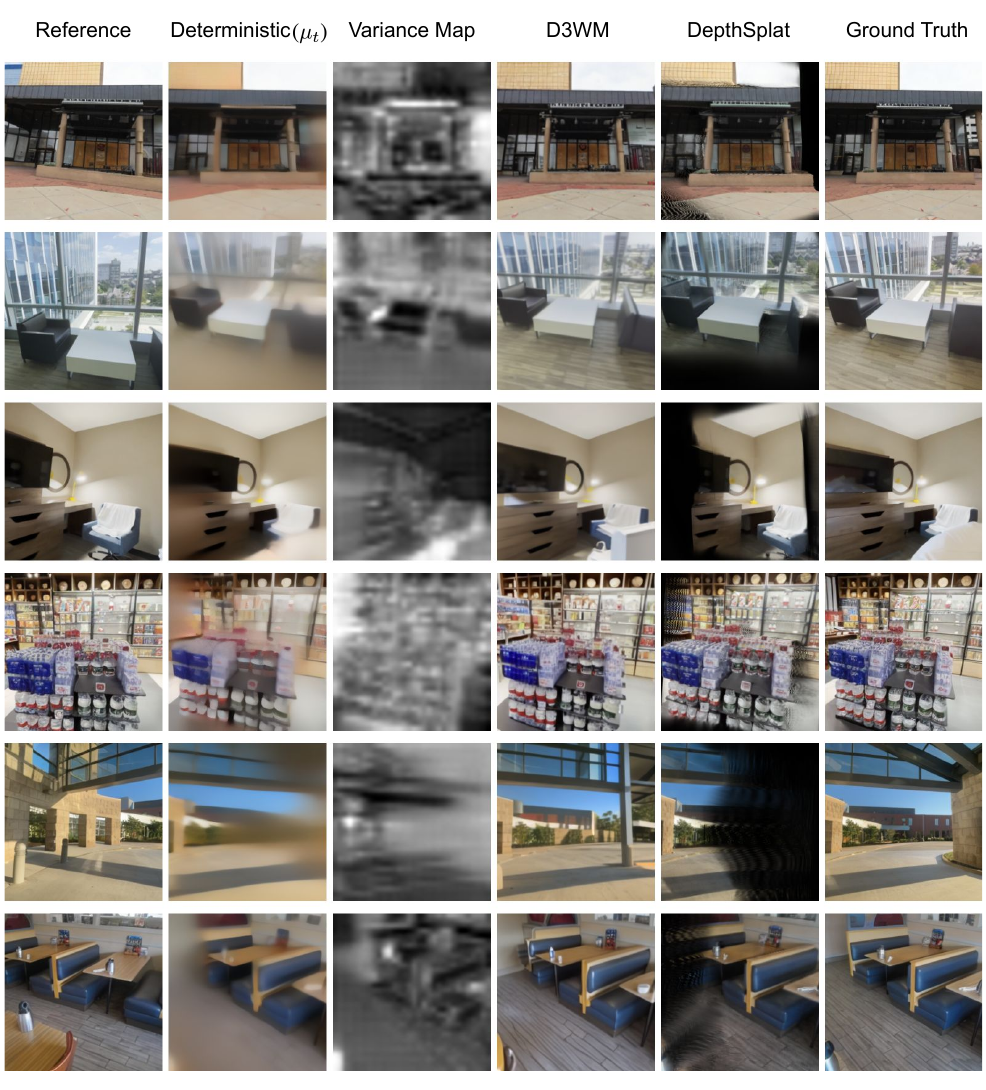}
    \caption{Additional results for qualitative comparison between D3WM and DepthSplat on DL3DV Benchmark dataset with extrapolation time shift $t=6$ frames into the future. Highlighting how D3WM handles the "black hole" problem as compared to 3DGS approaches while retaining structural consistency of the rest of scene.}
    \label{fig:dsdisp6_qual_supplement1}
\end{figure}

\begin{figure}[h!]
    \centering
     \includegraphics[width=\textwidth]{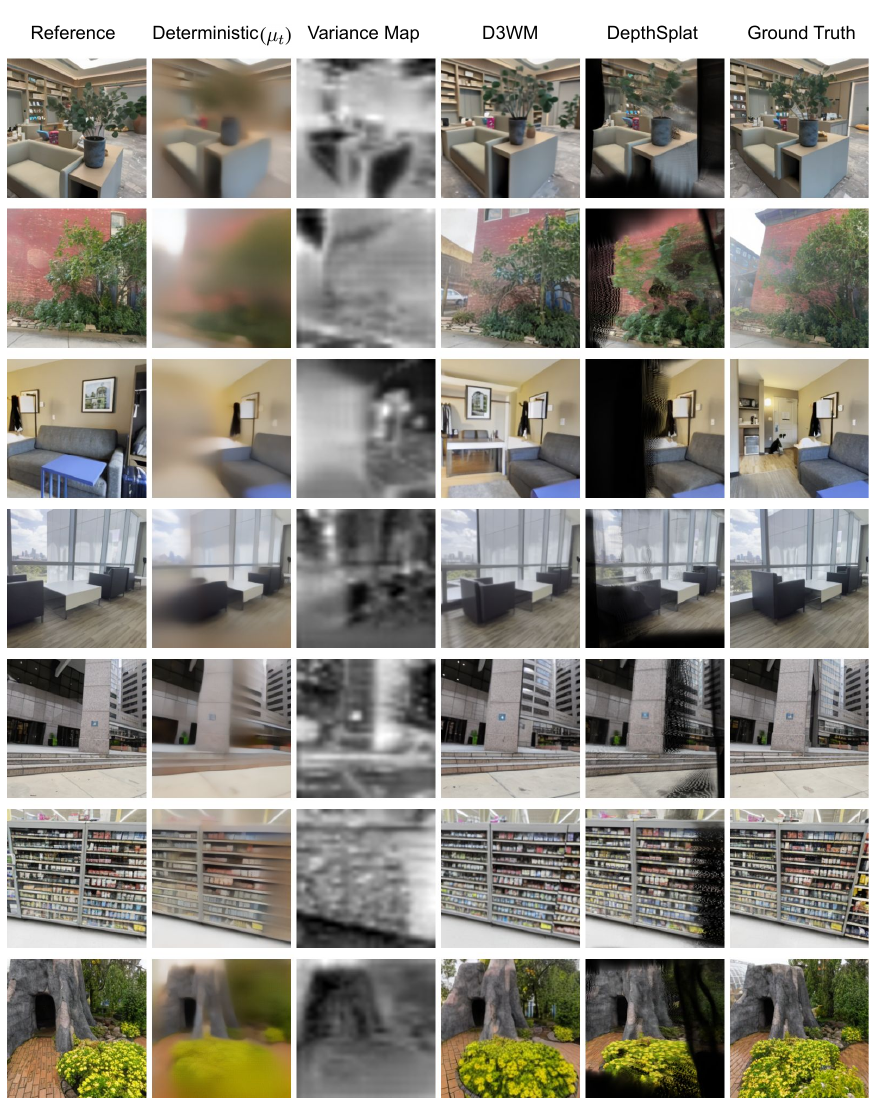}
    \caption{Additional results for qualitative comparison between D3WM and DepthSplat on DL3DV Benchmark dataset with extrapolation time shift $t=9$ frames into the future. Highlighting how D3WM handles the "black hole" problem as compared to 3DGS approaches while retaining structural consistency of the rest of scene.}
    \label{fig:dsdisp9_qual_supplement1}
\end{figure}

\begin{figure}[h!]
    \centering
     \includegraphics[width=\textwidth]{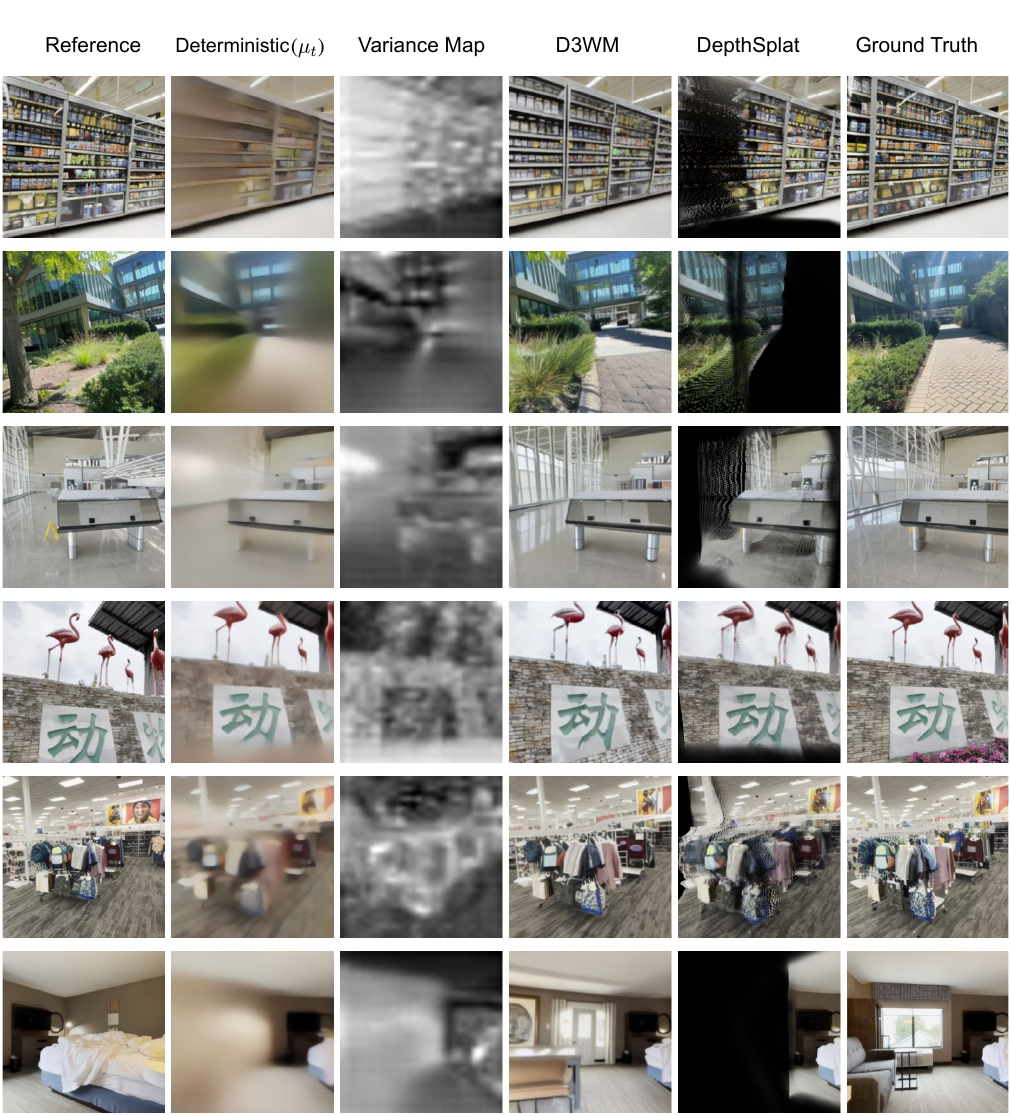}
    \caption{Additional results for qualitative comparison between D3WM and DepthSplat on DL3DV Benchmark dataset with extrapolation time shift $t=12$ frames into the future. Highlighting how D3WM handles the "black hole" problem as compared to 3DGS approaches while retaining structural consistency of the rest of scene.}
    \label{fig:dsdisp12_qual_supplement1}
\end{figure}

\section{NVS Results Discussion}\label{sec:nvs_discussion}

\textbf{PSNR-LPIPS-SSIM Across Horizons.} The DepthSplat comparison in
Table~\ref{tab:nvs_depthsplat_quant} shows that D3WM dominates on PSNR and LPIPS at 6-frame extrapolation and beyond, and on all three metrics at 9 and 12 frames. The persisting SSIM gap at the 3- and 6-frame horizons reflects a fundamental property of the methods rather than a weakness of D3WM: Gaussian splatting is a rendering technique that derives output pixels directly from input frames where geometric warping is sufficient, so when the field of view overlap is large (small extrapolation distance), splatting produces near-perfect pixel reconstruction in non-disoccluded regions and very high SSIM scores. As extrapolation distance grows, disocclusion volume grows with it; DepthSplat then leaves expanding ``black holes'' (Figures~\ref{fig:dsdisp3_qual_supplement1}, \ref{fig:dsdisp6_qual_supplement1}, \ref{fig:dsdisp9_qual_supplement1}, \ref{fig:dsdisp12_qual_supplement1}) where no input pixels can be projected. The squared-error penalty in PSNR strictly punishes these massive deviations, which is why DepthSplat's PSNR drops sharply with horizon while D3WM's diffusion network handles disoccluded regions cleanly.

\textbf{Deterministic Only vs Diffusion-Based Predictions.}
The ablation table (Section~\ref{sec:full_ablation_table}) shows that the \emph{Deterministic Only} baseline
(deterministic transformer alone) achieves the highest PSNR/SSIM scores but the second-worst LPIPS (only \emph{Gate Only} is worse).
This is the same fundamental metric tension affecting the DepthSplat comparison, but expressed within our own architecture. The deterministic prediction $\mu_t$ averages over all plausible outcomes in regions of high aleatoric uncertainty, producing locally-blurred predictions that preserve average pixel intensity. PSNR's squared-error penalty rewards this averaging because the per-pixel deviation from any single plausible outcome remains bounded, while LPIPS heavily penalizes the lack of high-frequency structural detail this blur produces. Our full method intentionally trades a small amount of PSNR/SSIM for significantly improved perceptual realism (LPIPS), with the sharp-but-occasionally-incorrect textures synthesized by diffusion in high-variance regions. The visualizations (Figures~\ref{fig:dsdisp3_qual_supplement1}, \ref{fig:dsdisp6_qual_supplement1}, \ref{fig:dsdisp9_qual_supplement1}, \ref{fig:dsdisp12_qual_supplement1}) make this tradeoff visible: deterministic-only outputs look identifiably blurry in disocclusion regions, while D3WM's outputs look natural.

\begin{figure}[h!]
    \centering
     \includegraphics[width=\textwidth]{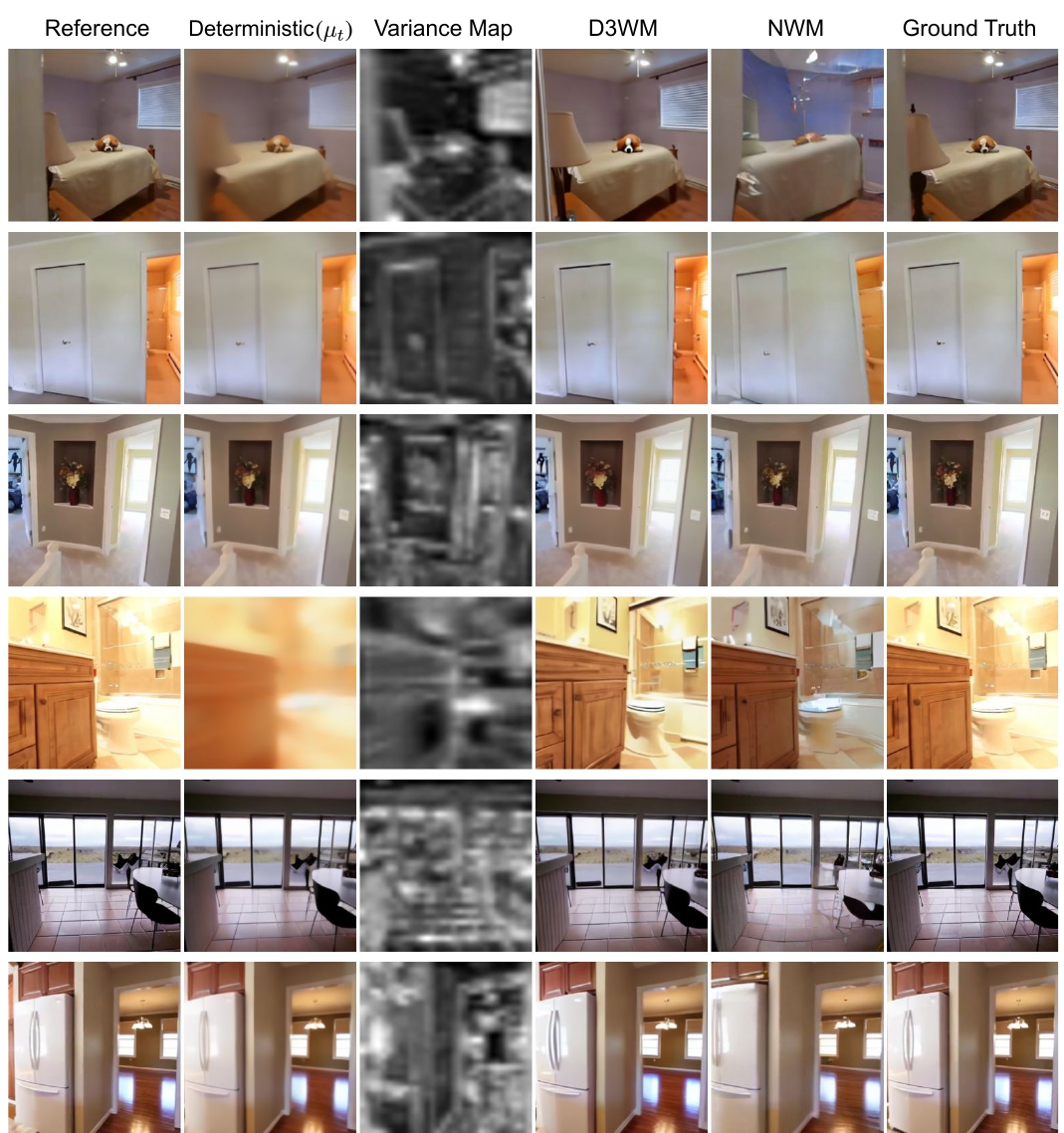}
    \caption{Additional results for qualitative comparison between D3WM and NWM on out of distribution RealEstate10K dataset subset with extrapolation time shift $t=3$ frames into the future. Highlighting how D3WM controls and handles the hallucinations that are common for diffusion-only approaches.}
    \label{fig:nwmdisp3_qual_supplement1}
\end{figure}

\begin{figure}[h!]
    \centering
     \includegraphics[width=\textwidth]{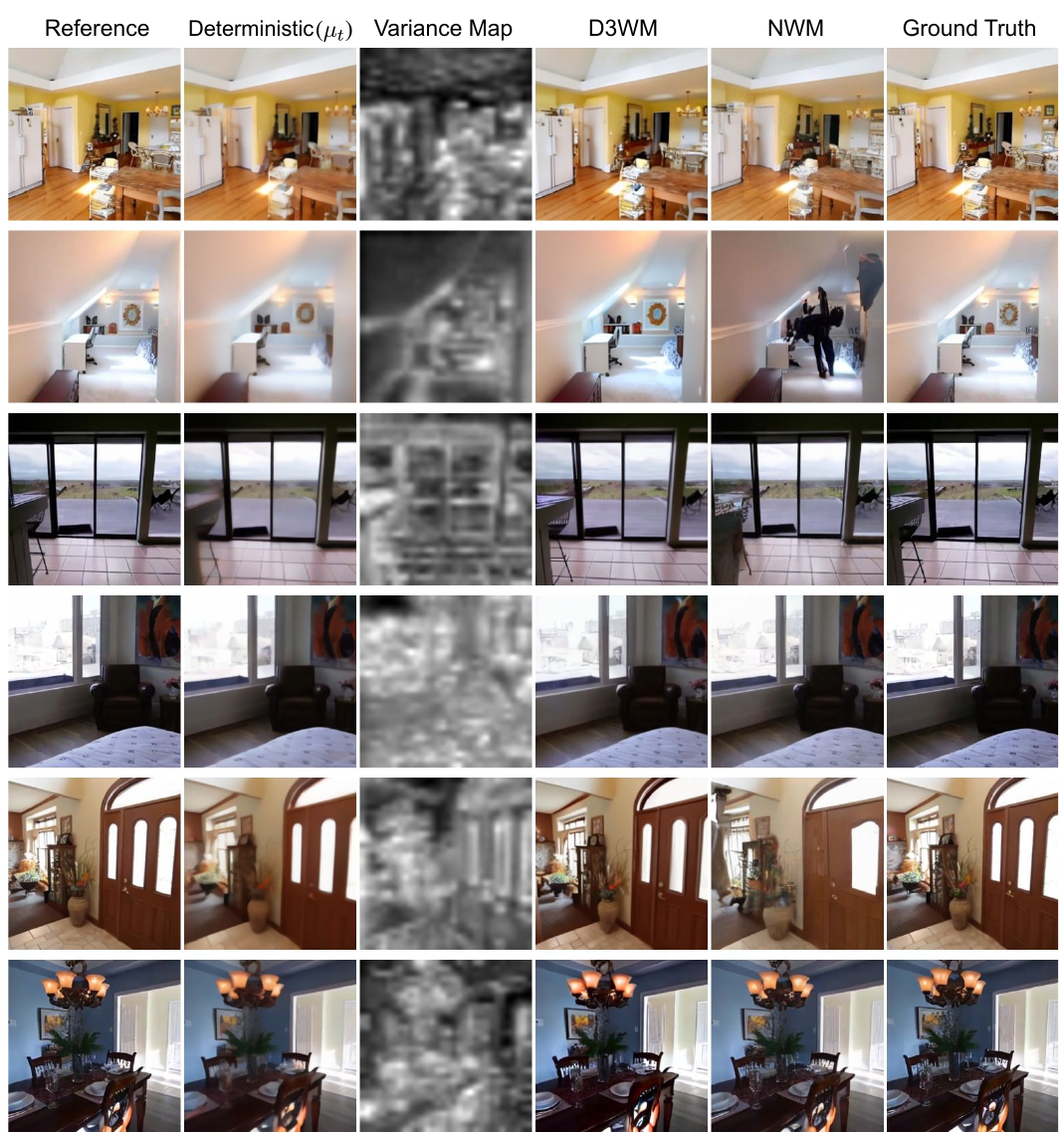}
    \caption{Additional results for qualitative comparison between D3WM and NWM on out of distribution RealEstate10K dataset subset with extrapolation time shift $t=6$ frames into the future. Highlighting how D3WM controls and handles the hallucinations that are common for diffusion-only approaches.}
    \label{fig:nwmdisp6_qual_supplement1}
\end{figure}

\begin{figure}[h!]
    \centering
     \includegraphics[width=\textwidth]{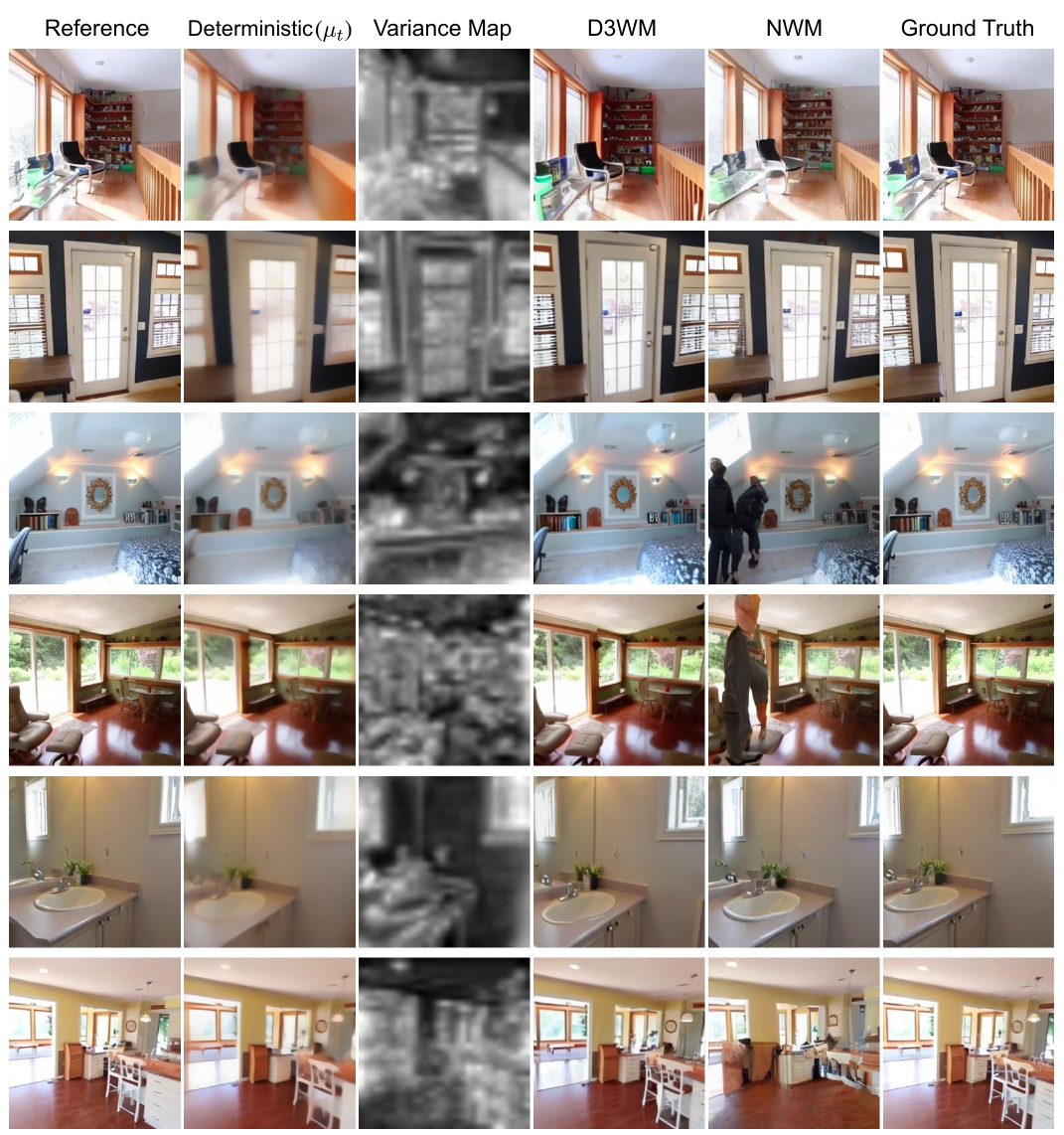}
    \caption{Additional results for qualitative comparison between D3WM and NWM on out of distribution RealEstate10K dataset subset with extrapolation time shift $t=9$ frames into the future. Highlighting how D3WM controls and handles the hallucinations that are common for diffusion-only approaches.}
    \label{fig:nwmdisp9_qual_supplement1}
\end{figure}

\begin{figure}[h!]
    \centering
     \includegraphics[width=\textwidth]{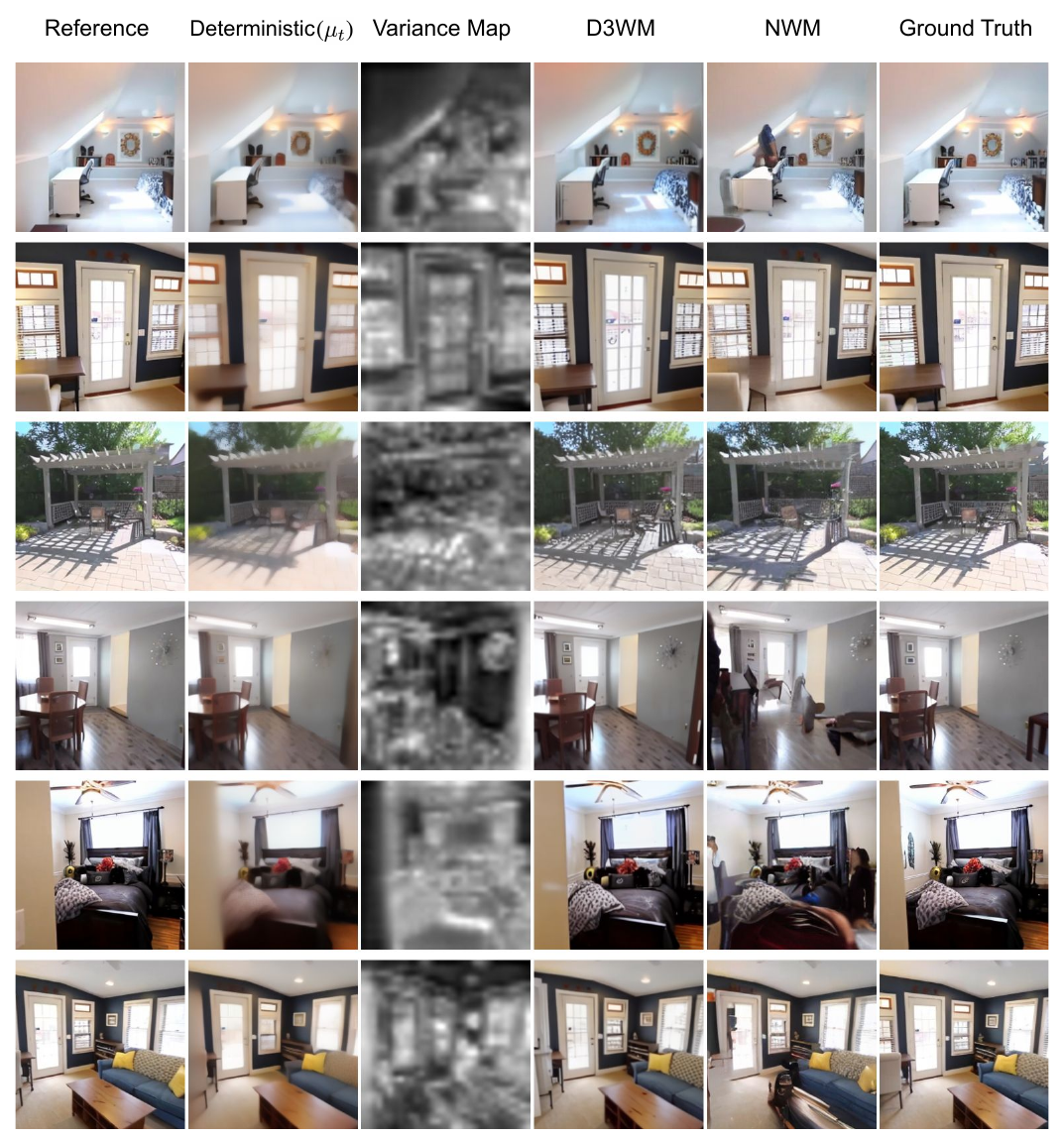}
    \caption{Additional results for qualitative comparison between D3WM and NWM on out of distribution RealEstate10K dataset subset with extrapolation time shift $t=12$ frames into the future. Highlighting how D3WM controls and handles the hallucinations that are common for diffusion-only approaches.}
    \label{fig:nwmdisp12_qual_supplement1}
\end{figure}

\textbf{Expanded Qualitative Comparison to NWM.}
D3WM produces a significant improvement in the visual consistency of generated frames compared to NWM. Figures \ref{fig:nwmdisp3_qual_supplement1},
\ref{fig:nwmdisp6_qual_supplement1}, \ref{fig:nwmdisp9_qual_supplement1},
and \ref{fig:nwmdisp12_qual_supplement1} show additional examples of
hallucinations and structure distortions produced by NWM compared to D3WM. These results support our motivation for grounding predictions by modeling deterministic processes first, and applying generative diffusion only in spatial areas of high stochasticity. The structural anchoring provided by the variance-gated bridge produces both higher PSNR and lower LPIPS than NWM's free-form diffusion approach across all four horizons, even though RealEstate10k was not used during training.

\begin{figure}[h!]
    \centering
     \includegraphics[width=\textwidth]{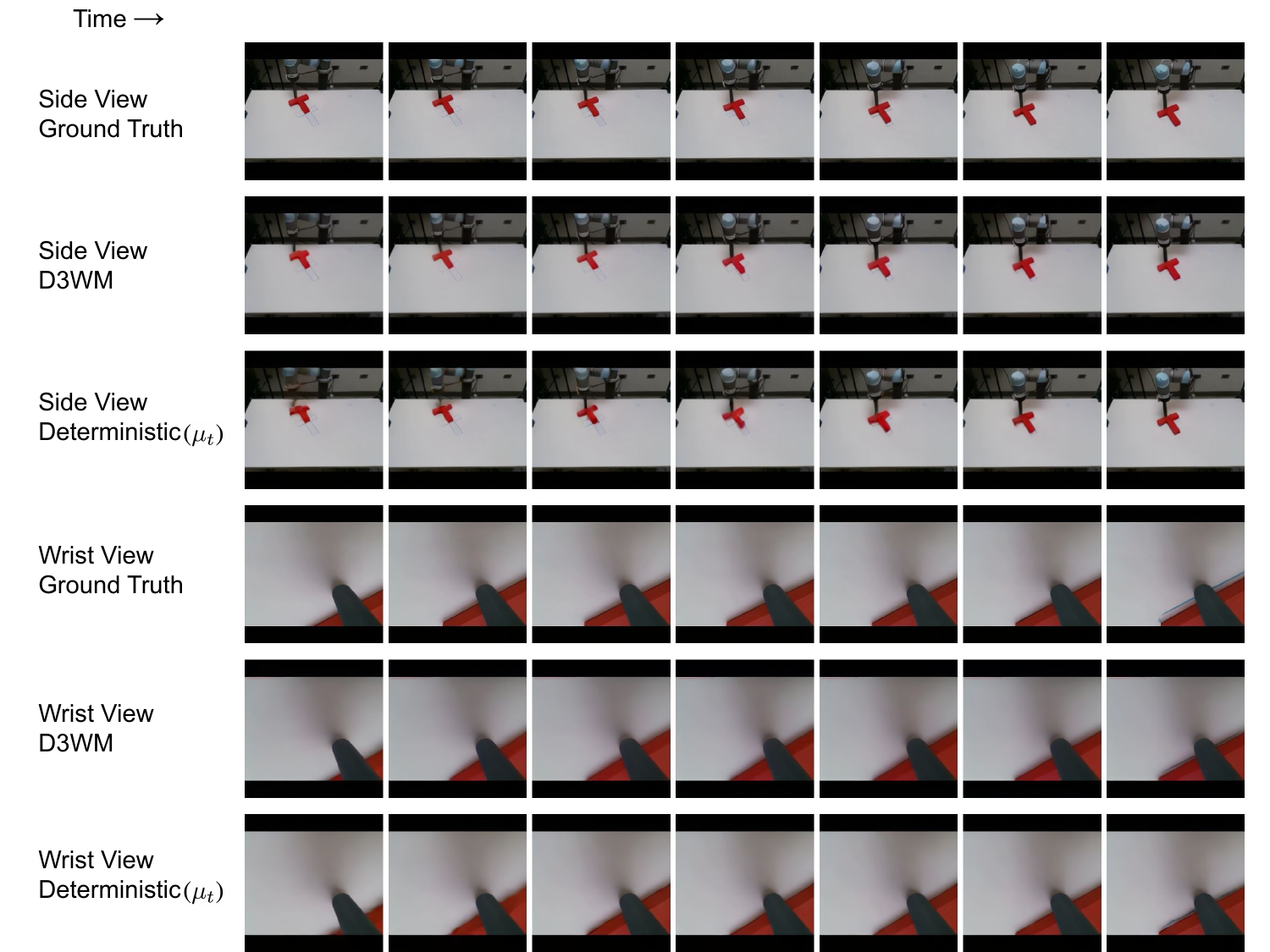}
    \caption{ Qualitative results on Real PushT test dataset. D3WM for PushT predicts both side view and wrist view, learning the correct correlation between the two and accurate point of contact with the object. In this example, we can see every 5th frame of predictions made in a row from ground truth frames, showing that we correctly capture the deterministic dynamics of the scene, as evident from the Deterministic predictions. These example frames show the object being pushed straight into the target location.
    }
    \label{fig:pusht_qual_supplement1}
\end{figure}

\begin{figure}[h!]
    \centering
     \includegraphics[width=\textwidth]{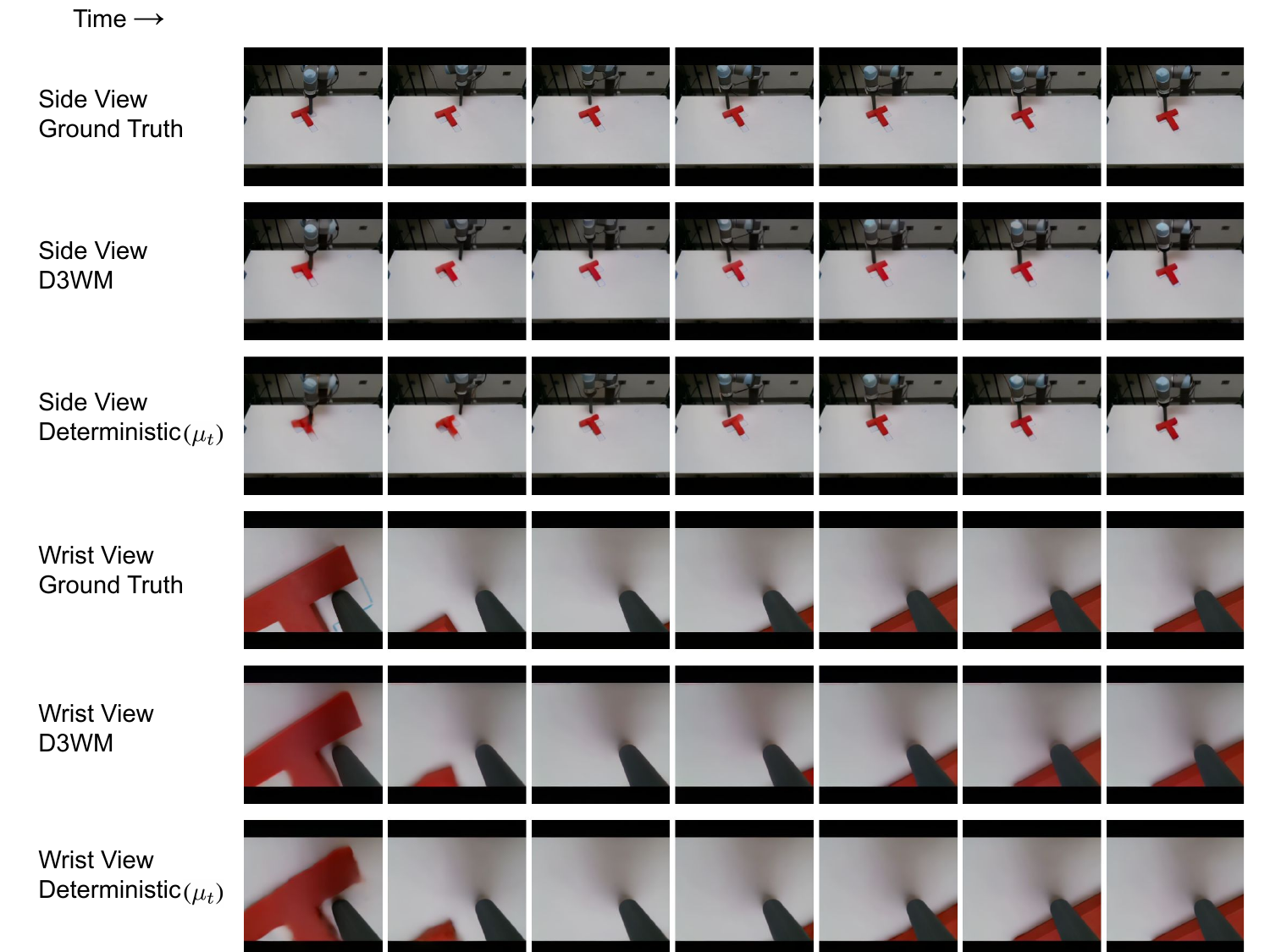}
    \caption{ Additional qualitative results on Real PushT test dataset. D3WM for PushT predicts both side view and wrist view, learning the correct correlation between the two and accurate point of contact with the object. In this example we can see every 5th frame of a predictions made in a row from ground truth frames, showing that we correctly capture the deterministic dynamics of the scene, as evident from the Drift Only predictions. The sample frames show the robotic gripper going around the object and then pushing it into the target location, with D3WM correctly capturing appearance and disappearance of the object in the wrist view.
    }
    \label{fig:pusht_qual_supplement2}
\end{figure}

\begin{figure}[h!]
    \centering
     \includegraphics[width=\textwidth]{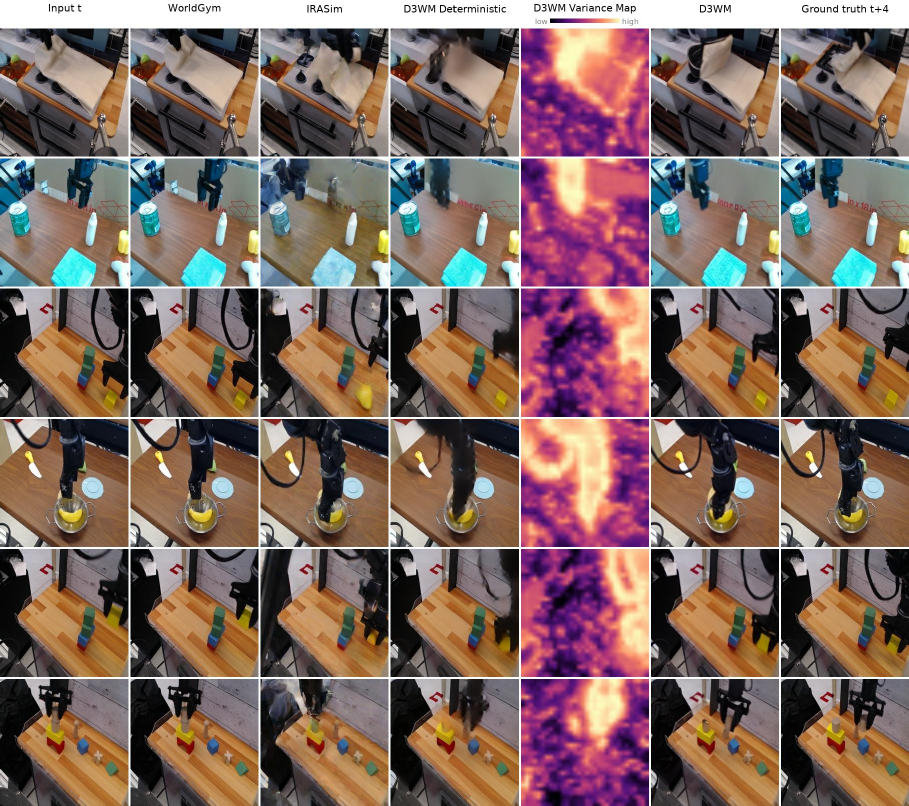}
    \caption{ Additional qualitative results on BridgeData~V2 dataset. D3WM Deterministic prediction correctly realizes the commanded motion, while blurring in the high-variance regions. Variance Map, plotted for visualization purposes only from its stretched latent representation, is used to sample a committed outcome using the stochastic component, resulting in a sharp D3WM prediction.}
    \label{fig:bridge_qual_supplement1}
\end{figure}

\begin{figure}[h!]
    \centering
     \includegraphics[width=\textwidth]{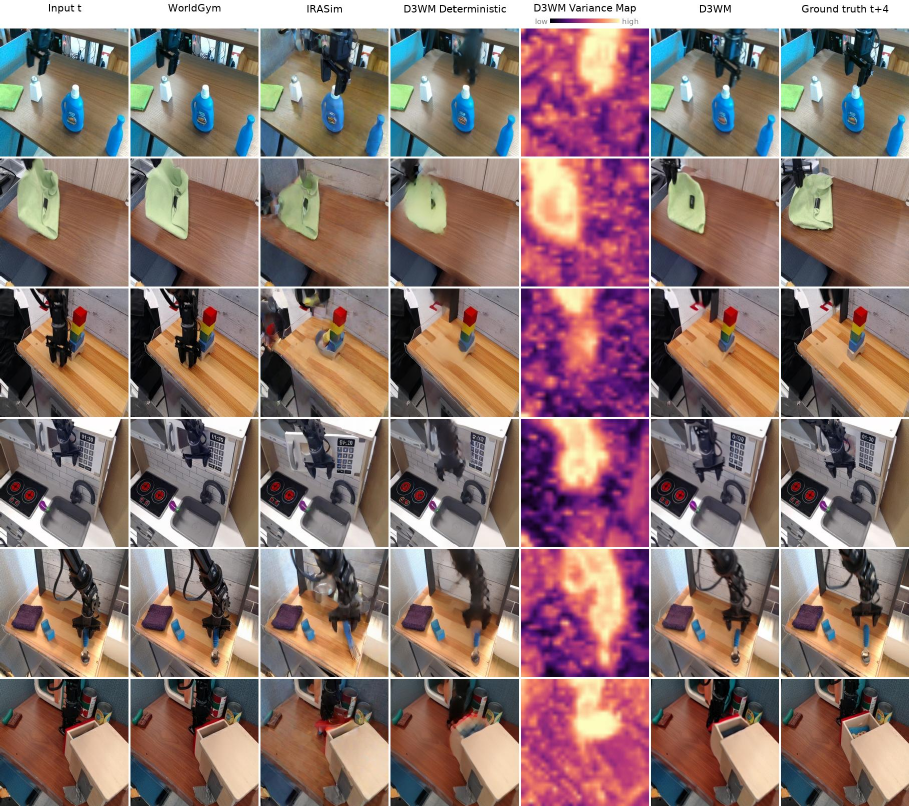}
    \caption{ Additional qualitative results on BridgeData~V2 dataset. D3WM Deterministic prediction correctly realizes the commanded motion, while blurring in the high-variance regions. Variance Map, plotted for visualization purposes only from its stretched latent representation, is used to sample a committed outcome using the stochastic component, resulting in a sharp D3WM prediction.}
    \label{fig:bridge_qual_supplement2}
\end{figure}

\section{PushT Results Discussion}

In the context of autonomous robots, particularly for precise manipulation tasks such as PushT, accurate prediction of the deterministic outcomes of actions is critical. Although Real PushT scenes naturally exhibit partial occlusions and inherent stochasticity, robust robotic planning fundamentally depends on the deterministic consequences of taken actions. Consequently, approaches that rely exclusively on autoregressive diffusion-based generation, such as iVideoGPT, frequently struggle in this domain. As demonstrated by the quantitative and qualitative results in Section~\ref{subsec:robot}, these purely generative models are highly susceptible to visual distortions and physically implausible hallucinations, which ultimately undermine downstream planning.

By contrast, as illustrated in Figures \ref{fig:pusht_qual_supplement1} and \ref{fig:pusht_qual_supplement2} , D3WM effectively overcomes these limitations by accurately modeling the deterministic outcomes of manipulation actions. This stability is largely driven by our deterministic image prediction ($\mu_t$)  and variance map, which anchors the generation process. Furthermore, the deterministic transformer within D3WM successfully anticipates the complex appearances and rigid-body movements of the "T" object from the wrist camera's perspective. By predicting this local wrist view concurrently with the side camera view, our model ensures that the resulting action-conditioned changes are actively informed by and highly consistent with the broader spatial context of the scene from different camera views.

\section{BridgeData~V2 Results Discussion}

One small difference in the architecture of the stochastic component for BridgeData~V2 is worth mentioning. Due to the mostly static background in the dataset, we found that injecting the reference frame into the ControlNet worked better via reference-attention rather than IP-Adapter. Additional qualitative results on BridgeData~V2 dataset can be found in Figures \ref{fig:bridge_qual_supplement1} and
\ref{fig:bridge_qual_supplement2}. Consistent with our observations, D3WM sometimes samples not the most precise stochastic outcomes in terms of end-effector shape and object colors and textures, but preserves the generally correct motion due to the deterministic prediction grounding. IRASim, on the other hand, frequently shows incorrect motion or distorts the objects' shapes and the background. WorldGym does generally much better in terms of consistency and sharpness, but realizes only a fraction of the commanded motion, and sometimes shows incorrect motion as well.

\section{Derivation of the Variance-Gated Inference Update}
\label{sec:sde_derivation}

In this section, we derive the variance-gated bridge inference of
Section~\ref{subsec:inference}. We work entirely in the Variance-Preserving (VP) regime that matches our underlying diffusion backbone (Stable Diffusion 1.5). We first present the noisy-bridge formulation used by D3WM, then contrast it with a clean-state posterior-mean alternative which we ablate in Appendix~\ref{sec:full_ablation_table} as \emph{Clean State Blend}.

Let $t$ denote the temporal frame index of the world model, and
$\tau \in [0, T]$ denote the diffusion noise level. We sample the next
state $z_t$ given the deterministic prediction $\mu_t$ and aleatoric
variance $\sigma_t^2$ produced by our transformer.

\subsection{Variance-Preserving Forward and Reverse SDE}

The VP forward diffusion process over $\tau$ is
\begin{equation}
    \mathrm{d} z_t^{(\tau)} = -\tfrac{1}{2}\beta(\tau)\, z_t^{(\tau)}\, \mathrm{d}\tau
        + \sqrt{\beta(\tau)}\, \mathrm{d}w_\tau,
    \label{eq:vp_forward}
\end{equation}
with closed-form marginal distribution
\begin{equation}
    z_t^{(\tau)} = \sqrt{\bar{\alpha}_\tau}\, z_t^{(0)}
        + \sqrt{1 - \bar{\alpha}_\tau}\, \epsilon,
    \quad \epsilon \sim \mathcal{N}(0, I),
    \label{eq:vp_marginal}
\end{equation}
where $\bar{\alpha}_\tau = \exp\!\big(\!-\!\!\int_0^\tau \beta(s)\,\mathrm{d}s\big)$
is the cumulative noise schedule. The corresponding reverse-time SDE that generates samples from noise is
\begin{equation}
    \mathrm{d} z_t^{(\tau)} = \Big[ -\tfrac{1}{2}\beta(\tau)\, z_t^{(\tau)}
        - \beta(\tau)\, \nabla_{z}\log p_\tau(z_t^{(\tau)}) \Big]\mathrm{d}\tau
        + \sqrt{\beta(\tau)}\, \mathrm{d}\bar{w}_\tau .
    \label{eq:vp_reverse}
\end{equation}
Tweedie's formula for the VP marginal gives the score in terms of the
conditional clean-state expectation,
\begin{equation}
    \nabla_z \log p_\tau(z_t^{(\tau)})
    = \frac{\sqrt{\bar{\alpha}_\tau}\, \mathbb{E}[z_t^{(0)}\mid z_t^{(\tau)}]
        - z_t^{(\tau)}}{1 - \bar{\alpha}_\tau} .
    \label{eq:tweedie_vp}
\end{equation}
We omit the $t$-subscript on the score function for clarity; all quantities pertain to the world-model timestep $t$ being predicted.

\subsection{Deterministic Prior from the Transformer Network}

The deterministic prediction network $f_\theta^\mu$ provides a Gaussian prior over the clean target state given the input observation $z_0$ and action sequence $a_{[0,t)}$:
\begin{equation}
    p(z_t^{(0)}\mid z_0, a_{[0,t)}) = \mathcal{N}\!\big(z_t^{(0)};\,
        \mu_t,\ \sigma_t^2 \odot I\big),
    \label{eq:drift_prior}
\end{equation}
with spatial diagonal variance $\sigma_t^2 \in [e^{-10}, 1]^{C \times h \times w}$ (Section~\ref{subsec:drift}). The gate, the attention bias and the ControlNet conditioning use its channel average, an $h \times w$ map broadcast across channels (in Eqs.~\ref{eq:trust_map_supp}, \ref{eq:bridge_blend}, $\sigma_t^2$ denotes this average).
This prior captures the heteroscedastic aleatoric uncertainty learned by $f_\theta^\mu$: low variance in regions where the action's effect is deterministic (e.g., visible structures undergoing predictable warping) and high variance in regions where it is stochastic (e.g., disocclusions).

\subsection{Noisy-Bridge Inference (D3WM Main Method)}
\label{subsec:noisy_bridge}

Pure unconditional diffusion samples from the unconditional latent prior $p(z_t^{(0)})$. Pure deterministic prediction returns $\mu_t$. We construct an intermediate process that respects the deterministic prior strongly where $\sigma_t^2$ is small, while leaving the diffusion model free to act where $\sigma_t^2$ is large.

\paragraph{Spatially-gated forward bridge.} We define the per-pixel trust map
\begin{equation}
    \gamma_t \;=\; (\mathbf{1} - \sigma_t^2) \odot \lambda_\tau,
    \label{eq:trust_map_supp}
\end{equation}
where $\lambda_\tau \in [0, 1]$ is a diffusion-time-dependent anchoring schedule controlling the overall strength of deterministic anchoring. At each reverse step, we replace the diffusion model's noisy latent $z_t^{(\tau)}$ with a per-pixel convex combination
\begin{equation}
    \tilde{z}_t^{(\tau)} \;=\; \gamma_t \odot \tilde{\mu}_t^{(\tau)}
        + (\mathbf{1} - \gamma_t) \odot z_t^{(\tau)},
    \label{eq:bridge_blend}
\end{equation}
where $\tilde{\mu}_t^{(\tau)}$ is the deterministic prior mean transported to the noise level of step $\tau$ via the same VP forward marginal as the data,
\begin{equation}
    \tilde{\mu}_t^{(\tau)} \;=\; \sqrt{\bar{\alpha}_\tau}\, \mu_t
        + \sqrt{1 - \bar{\alpha}_\tau}\, \xi_\tau,
        \qquad \xi_\tau \sim \mathcal{N}(0, I).
    \label{eq:noised_drift}
\end{equation}
Because both terms in the convex combination are valid noisy latents at level $\tau$ (Eq.~\ref{eq:vp_marginal}), the blended latent
$\tilde{z}_t^{(\tau)}$ also lies on the noise-level manifold expected by the next reverse step.
We then take a standard DDIM step (a discretization of the probability-flow ODE associated with Eq.~\ref{eq:vp_reverse})
using $\tilde{z}_t^{(\tau)}$ in place of $z_t^{(\tau)}$.

\paragraph{Limit behavior.} Eq.~\ref{eq:bridge_blend} realizes the
qualitative behavior described in the main paper:
\begin{itemize}
    \item \textbf{Low aleatoric uncertainty} ($\sigma_t^2 \!\to\! 0$):
    $\gamma_t \!\to\! \lambda_\tau$. With $\lambda_\tau$ near $1$, the
    blended latent is dominated by the noisy deterministic prior
    $\tilde{\mu}_t^{(\tau)}$, and the final $\hat{z}_t$ collapses toward $\mu_t$. Geometric structure is preserved.
    \item \textbf{High aleatoric uncertainty} ($\sigma_t^2 \!\to\!1$): $\gamma_t \!\to\! 0$. The blended latent equals
    $z_t^{(\tau)}$, and the diffusion model freely synthesizes content.
    \item \textbf{Schedule effect} (small $\lambda_\tau$): late in denoising the schedule reduces the deterministic anchoring even in low-variance regions, allowing the diffusion model to refine fine textures. In our default configuration we use a step-function schedule that disables anchoring after the first 70\% of denoising steps (Appendix~\ref{sec:full_ablation_table}).
\end{itemize}

\subsection{Alternative: Clean-State Posterior-Mean Blending}
\label{subsec:clean_state}

For comparison and motivation of our design choice, we derive the
alternative \emph{clean-state} formulation that we ablate in Appendix~\ref{sec:full_ablation_table}. In this alternative, the deterministic prior is incorporated as a posterior correction to the diffusion model's clean-state estimate at each step, rather than as a noisy-latent replacement.

At each step $\tau$, the diffusion model produces a clean-state estimate
$\hat{z}_{t,\tau}^{(0)}$ via the standard VP $\epsilon$-prediction
\begin{equation}
    \hat{z}_{t,\tau}^{(0)} \;=\;
    \frac{z_t^{(\tau)} - \sqrt{1 - \bar{\alpha}_\tau}\, \hat{\epsilon}}
         {\sqrt{\bar{\alpha}_\tau}},
\end{equation}
with implicit variance
\begin{equation}
    \rho_\tau^2 \;=\; \frac{1 - \bar{\alpha}_\tau}{\bar{\alpha}_\tau},
    \label{eq:rho_tau}
\end{equation}
which is the variance of $\hat{z}_{t,\tau}^{(0)}$ around the true
$z_t^{(0)}$ under the noise schedule.

We have two Gaussian sources of information about $z_t^{(0)}$:
\begin{itemize}
    \item The temporal prior
    $\mathcal{N}(\mu_t, \sigma_t^2 I)$ from the deterministic network.
    \item The diffusion observation $\mathcal{N}(\hat{z}_{t,\tau}^{(0)}, \rho_\tau^2 I)$ from the score model.
\end{itemize}
The Gaussian posterior over $z_t^{(0)}$ has a precision-weighted mean
\begin{equation}
    \tilde{z}_{t,\tau}^{(0)} \;=\;
    \frac{\sigma_t^{-2} \odot \mu_t + \rho_\tau^{-2} \odot \hat{z}_{t,\tau}^{(0)}}
         {\sigma_t^{-2} + \rho_\tau^{-2}}.
\end{equation}
Multiplying numerator and denominator by $\sigma_t^2 \odot \rho_\tau^2$,
\begin{equation}
    \tilde{z}_{t,\tau}^{(0)} \;=\;
    \frac{\rho_\tau^2}{\rho_\tau^2 + \sigma_t^2} \odot \mu_t
    + \frac{\sigma_t^2}{\rho_\tau^2 + \sigma_t^2} \odot \hat{z}_{t,\tau}^{(0)}.
    \label{eq:clean_state_blend}
\end{equation}
Defining $\gamma_t^{\text{cs}} = \rho_\tau^2 / (\rho_\tau^2 + \sigma_t^2)$ gives the same convex-combination form as the noisy-bridge blend (Eq.~\ref{eq:bridge_blend}), with the clean quantities $\mu_t$ and $\hat{z}_{t,\tau}^{(0)}$ in place of the noisy latents $\tilde{\mu}_t^{(\tau)}$ and $z_t^{(\tau)}$, and a weight set by the noise level through $\rho_\tau^2$ rather than by the schedule $\lambda_\tau$.

\paragraph{Why the noisy-bridge outperforms.} The clean-state formulation in Eq.~\ref{eq:clean_state_blend} has a structural failure mode that explains its inferior empirical performance (see \emph{Clean State Blend} ablation in Appendix~\ref{sec:full_ablation_table}). At early diffusion steps $\tau$, the implicit variance $\rho_\tau^2 = (1 - \bar{\alpha}_\tau)/\bar{\alpha}_\tau$ of $\hat{z}_{t,\tau}^{(0)}$ becomes very large because $\bar{\alpha}_\tau$ approaches zero. Consequently, the precision-weighted average gives disproportionate weight to $\mu_t$ in early denoising steps everywhere in the image, regardless of the spatial variance map. This over-anchors the trajectory to the blurry deterministic prediction in early steps, leaving the diffusion model insufficient room across the remaining steps to recover sharp textures even in high-uncertainty regions.

The noisy-bridge formulation
(Section~\ref{subsec:noisy_bridge}) avoids this failure because the
schedule $\lambda_\tau$ that controls overall anchoring strength is
independent of the diffusion noise level — we can keep $\lambda_\tau$
near $1$ for the first portion of denoising (where the diffusion model
benefits from the deterministic prior's structure) and reduce it to $0$ for the second portion (where the diffusion model produces texture
detail). The spatial gate $(\mathbf{1} - \sigma_t^2)$ and the temporal
schedule $\lambda_\tau$ are decoupled.

\subsection{Relationship to the Temporal SDE}

The world-model temporal SDE in Eq.~\ref{eq:temporal_sde} models the
\emph{data-generating process} along the trajectory axis $t$. The
deterministic prediction network regresses the integrated effect of the temporal drift coefficient over $[0, t]$ (i.e., the conditional mean $\mu_t = \mathbb{E}[z_t \mid z_0, a_{[0,t)}]$) together with the spatial variance $\sigma_t^2$ of the unobserved residual. We do not numerically integrate the temporal SDE, instead, the trained network $f_\theta^\mu$ produces the integrated mean and variance directly. The reverse-time diffusion described in this section operates on a separate axis $\tau$ (diffusion noise level) and provides image-manifold-aware sampling from the conditional distribution $p(z_t \mid z_0, a_{[0,t)})$. The two SDEs share only the spatial variance map $\sigma_t^2$, which couples them via the trust map $\gamma_t$.

\section{Model Architecture Details}\label{sec:arch}

We built our deterministic prediction network $f_\theta^\mu$ usinga vision transformer architecture with about 450 million parameters, also using AdaLN for conditioning. We used an MLP to embed the action conditioning so that it fits up to $7 \times 16$ size, the size requirement dictated by our PushT experiment. This size can be easily changed for action spaces with different dimensions.
The NVS deterministic prediction network outputs two latent maps, the predicted mean $\mu_t$ and the variance map $\sigma_t^2$.
Our PushT network outputs four images in latent space:
the predicted mean and variance map for the side camera view, and the same pair for the wrist camera view.
For the NVS task, our deterministic network takes two reference frames as input. For the PushT task, our deterministic network takes one wrist view and one side view as input.

We used the Stable Diffusion Canny ControlNet architecture to build the diffusion component of D3WM. We used Stable Diffusion 1.5 together with the IP Adapter Plus for Stable Diffusion 1.5 as the backbone. We changed the control input layer of our ControlNet to condition on the latent mean prediction $\mu_t$ and the variance map. During inference, all of the gating and selective noising of the intermediate frames during each diffusion step are done in latent space using the latent deterministic prediction image ($\mu_t$) and the latent variance map ($\sigma^2_t$) %
output from the deterministic network $f_\theta^\mu$. The variance map also biases the UNet self-attention: the attention logits of each query location are divided by a temperature $1 + \alpha\,\bar{\sigma}_t^2$ with $\alpha = 2$, where $\bar{\sigma}_t^2$ is the channel-averaged variance map resized to the attention resolution, so uncertain locations attend more broadly.

\textbf{BridgeData~V2.} The BridgeData~V2 model uses the same deterministic prediction network as the NVS model and, like it, outputs two latents, the predicted mean $\mu_t$ and the variance map $\sigma_t^2$, for the single fixed camera view. The action sequence is embedded with the same MLP and the same $7 \times 16$ input size as in PushT, here holding up to 16 steps of 7-dimensional actions (6-DoF end-effector motion plus gripper). The diffusion component uses the same ControlNet architecture as the other experiments, except that the reference frame is injected through reference attention rather than IP-Adapter, which worked better on the mostly static BridgeData~V2 scenes.

\section{Compute Resources}

We trained all models on a workstation with two NVIDIA RTX PRO 6000 Blackwell GPUs (96 GB GDDR7 each). Training the deterministic prediction network for NVS required approximately one week of wall-clock time across both GPUs (1M training steps at global batch size 256). The PushT deterministic prediction network trained considerably faster, in approximately 12 hours, due to a smaller batch size (32) and 5x fewer training steps (200k) for the simpler scene distribution. The conditional diffusion ControlNet required approximately 2 days for NVS (100k steps at batch size 16, with conditioning images at 512x512 resolution) and approximately 10 hours for PushT (20k steps at batch size 16). For BridgeData~V2, the deterministic prediction network trained for 400k steps, with additional training to 750k steps yielding diminishing returns, and the ControlNet for 100k steps. We note that comparable performance was achieved by the NVS deterministic network at 300k training steps with additional LPIPS and SSIM auxiliary loss terms, and the additional 700k steps yielded diminishing returns.

\end{document}